\documentclass{article}
\usepackage{amsmath,amssymb}
\usepackage{subcaption}
\usepackage{bm}
\usepackage{graphicx}
\usepackage{natbib}

\usepackage[preprint]{neurips_2021}

\usepackage[utf8]{inputenc} % allow utf-8 input
\usepackage[T1]{fontenc}    % use 8-bit T1 fonts
\usepackage[hidelinks]{hyperref}       % hyperlinks
\usepackage{url}            % simple URL typesetting
\usepackage{booktabs}       % professional-quality tables
\usepackage{amsfonts}       % blackboard math symbols
\usepackage{nicefrac}       % compact symbols for 1/2, etc.
\usepackage{microtype}      % microtypography
\usepackage{xcolor}         % colors
\usepackage{bm}
\usepackage{amsthm, amsmath}
\usepackage{pifont}
\usepackage[normalem]{ulem}

\newcommand{\integers}{{\mathbb{Z}}}

\newcommand{\EE}{\mathcal{E}}
\newcommand{\GG}{\mathcal{G}} 
\newcommand{\VV}{\mathcal{V}} 
\newcommand{\NN}{\mathcal{N}} 
 
\renewcommand{\AA}{\mathcal{A}} 
\newcommand{\R}{\mathbb{R}}
\newcommand{\Z}{\mathbb{Z}}

\newcommand{\E}{\mathbb{E}}

\newcommand{\qq}{\mathbf{q}}
\newcommand{\vv}{\mathbf{v}}
\newcommand{\ee}{\mathbf{e}}
\newcommand{\xx}{\mathbf{x}}
\newcommand{\yy}{\mathbf{y}}
\newcommand{\uu}{\mathbf{u}}
\newcommand{\zz}{\mathbf{z}}

\renewcommand{\aa}{\bm{\alpha}}
 
\newcommand{\CO}{\text{CO}}

\newcommand{\ie}{\emph{i.e.}}
\newcommand{\eg}{\emph{e.g.}}

\newcommand*\samethanks[1][\value{footnote}]{\footnotemark[#1]}

\newtheorem*{definition}{Definition}%[section]

\title{Symmetry-driven graph neural networks}

\author{%
Francesco Farina \thanks{Equal contribution} \\
  GSK.ai\\
  GlaxoSmithKline, London \\
  \texttt{francesco.x.farina@gsk.com}
   \And
  Emma Slade \samethanks \\
 
  GSK.ai\\
  GlaxoSmithKline, London \\ 
  \texttt{emma.x.slade@gsk.com}

}

\begin{document}
\maketitle

\begin{abstract}
Exploiting symmetries and invariance in data is a powerful, yet not
fully exploited, way to achieve better generalisation with more
efficiency.  In this paper, we introduce two graph network
architectures that are equivariant to several types of transformations
affecting the node coordinates.  First, we build equivariance to any
transformation in the coordinate embeddings that preserves the
distance between neighbouring nodes, allowing for equivariance to the
Euclidean group. Then, we introduce angle attributes to build
equivariance to any angle preserving transformation - thus, to the
conformal group.  Thanks to their equivariance properties, the
proposed models can be vastly more data efficient with respect to
classical graph architectures, intrinsically equipped with a better
inductive bias and better at generalising.  We demonstrate these
capabilities on a synthetic dataset composed of $n$-dimensional
geometric objects.  Additionally, we provide examples of their
limitations when (the right) symmetries are not present in the data.
\end{abstract}

\section{Introduction}
Symmetries exist throughout nature. All the fundamental laws of
physics are built upon the framework of symmetries, from the gauge
groups describing the Standard Model of particle physics, to
Einstein's theories of general and special relativity. Once one
understands the symmetry of a certain system, powerful predictions can
be made. A notable example is that of Gell-Mann's
eightfold-way~\citep{osti_4008239}, built upon the symmetries observed
in hadrons, that led to his prediction of the $\Omega^-$ baryon, which
was subsequently observed 3 years later~\citep{1964PhRvL..12..204B}.
The study of symmetries and invariance in deep learning has recently
become a field of interest to the community (see,
\eg,~\citep{bronstein2021geometric} for a comprehensive overview), and
rapid progress has been made in constructing architectures with group
theoretic structures embedded within. Two fundamental architectures in
machine learning, the convolutional and graph neural networks, are
invariant to the translation and symmetry groups respectively.

Graph networks in particular are designed to learn from
graph-structured data and are by construction permutation equivariant
with respect to the input.  They were originally proposed
in~\citep{gori2005new,scarselli2008graph} and have received a lot of
attention in the last years (see,
\eg,~\citep{battaglia2018relational,hamilton2020graph,wu2020comprehensive}
for a comprehensive overview). Due to their properties, they find
application in a broad range of problems like learning the dynamics of
complex physical systems~\citep{sanchez2020learning,
  pfaff2021learning}, particle identification in particle
physics~\citep{dreyer2021jet}, learning causal and relational
graphs~\citep{kipf2018neural, li2020causal}, discovering symbolic
models~\citep{cranmer2020discovering}, as well as quantum
chemistry~\citep{gilmer2017neural} and drug
discovery~\citep{STOKES2020688}.

In this work, we start from standard graph neural networks and extend
them to build, in addition to permutation equivariance, equivariance
to many transformations affecting the node coordinates, possibly
belonging to important groups.
First, we define the \emph{distance preserving graph network} (DGN)
which, thanks to modified update rules, is equivariant to any
transformation in the coordinate embeddings that preserves the
distance between neighbouring nodes. An example of such a
transformation is the Hoberman sphere, whose 3D shape changes while
preserving distances between connected nodes. As an important
particular case, we show that the architecture is equivariant to the
$n$-dimensional Euclidean group. An additional input layer can also
make it equivariant to dilations in the coordinates via the conformal
orthogonal group.
Then, we define the \emph{angle preserving graph network} (AGN), which
is equivariant to any transformation preserving the angles between
triples of neighbouring nodes in the graph (molecular conformations
are a notable example of such transformations) and, thus, to the
$n$-dimensional conformal group on $\R^n$.

By constructing such architectures, we enable a wide range of possible
transformations to be performed on graph-structured data, with the
only requirement being that distances or angles between coordinates of
neighbouring nodes are preserved. This means that both networks are
equivariant to translations, rotations, reflections (\ie, the
Euclidean group, $E(n)$), with the AGN being also equivariant to
dilations, inversions and non-orthogonal rotations.
In practice, this means the networks are able to filter out many
copies of the same input whose coordinate embeddings have been simply
rotated, moved or scaled and therefore learn more efficiently than
architectures which consider the inputs as distinct. In other words, a
single sample contains the same information as many copies of it
obtained by appropriately transforming it.
Finally, while the two architectures we present are \emph{partially}
overlapping in terms of equivariance features, it is important to
consider specific use cases where the use of one architecture over
another would be preferred.

Thanks to their properties, the proposed models are able to learn much
more efficiently and generalise better than standard graph networks,
even when data are transformed in highly non-trivial ways.
We demonstrate this on a synthetic dataset consisting of
$n$-dimensional geometric shapes and other benchmark datasets.
Moreover, we show examples of cases in which using our architectures
is counterproductive due to the lack of symmetries in the data.

\section{Background}
\subsection{Group theory and equivariance} \label{sec:groups}
 Group theory provides the mathematical formulation for symmetries of
 systems, with symmetry operations represented by individual group
 elements.
A group can be defined as a set $G$ equipped with a binary operation,
which enables one to combine two group elements to form a third,
whilst preserving the group axioms (associativity, identity, closure
and inverse).
A function unaffected by a group action is said to be
\emph{invariant}, which is particular instance of \emph{equivariance}.
Formally, let $X\subseteq \R^n$ and $\varphi_g :X\to X$ be a transformation on $X$ for a group element $g \in G$. Then, the linear map $\Phi:X\to Y$, $Y\subseteq \R^n$, is equivariant
to $G$ if $\forall g \in G$, $\exists\varphi_g ':Y\to Y$ such that
\begin{equation*}
\Phi(\varphi_g(\xx)) = \varphi_g'(\Phi(\xx))\,,\quad  \forall \varphi_g :X\to X\,.
\end{equation*}
When $\varphi_g '$ is the identity, we say that $\Phi$ is invariant to
$G$. Clearly, the functional form of $\phi$ is determined by the specific group of interest.
Next we introduce the Euclidean, $E(n)$, and the
conformal group.

It is worth pointing out that, whilst not often expressed in group
theoretic notation, equivariances exist in common deep learning
architectures; convolutional neural networks are equivariant under the
translation group $T(n)$, whilst graph neural networks are equivariant
to the symmetric group $S_n$.

\subsubsection{$E(n)$ equivariance} \label{sec:construct_en_equiv}
The translation and orthogonal groups ($T(n)$ and $O(n)$ respectively)
act through translations, orthogonal rotations and reflections. The semidirect
product of these two groups $E(n) = T(n) \rtimes O(n)$ is known as the
Euclidean group.
A transformation $\varphi$ is said to be $E(n)$ equivariant if it
satisfies the mapping $\varphi: \E^n \to \E^n$, where $\E^n$ is a
Euclidean space which can be thought of $\R^n$ with an inner product (for example, the dot product).
A function that is equivariant under $E(n)$ is $\|\xx_i-\xx_j\|_2^2$,
for any $\xx_i,\xx_j\in\E^n$ (see Appendix~\ref{sec:En_eq} for a
proof).

\subsubsection{Conformal equivariance} \label{sec:conformal_intro}

The conformal group $\text{Conf}(\R^{n,0})$ is the group of transformations $\varphi : \R^n \to \R^n$ that preserve angles. Formally, for any triple of vectors $\xx_j,\xx_i,\xx_k\in\R^n$, let us denote by $\angle (\xx_j,\xx_i,\xx_k)$ the angle
centred on $\xx_i$ with rays given by $\xx_j-\xx_i$ and $\xx_k-\xx_i$. Then, a conformal
transformation $\varphi$ satisfies
\begin{equation}\label{eq:conf_eq}
\angle (\xx_j,\xx_i,\xx_k) \to \angle (\varphi (\xx_j), \varphi(\xx_i), \varphi ( \xx_k) )=\angle (\xx_j, \xx_i, \xx_k) \ \,.
\end{equation}
By definition, the transformations of the
conformal group include translations, rotations reflections (which
collectively form the Euclidean group $E(n)$, as well as dilations,
inversions and other features.
There may be instances where one does not want to be equivariant under
the full conformal group as the broad range of allowed transformations
on the data may be a hinder. An important subgroup, which we consider,
known as the conformal orthogonal group requires transformations to be
orthogonal, and therefore inversions are not allowed.
If we construct our coordinate space $\xx$ to be equivariant under an
orthogonal transformation as well as a suitable dilation $\xx \to
\gamma \xx$, then we naturally will obtain an architecture which is
equivariant under the conformal orthogonal group as we show in
Appendix~\ref{sec:CO}. In other words, with a suitable definition of
$\gamma$ and $\xx$, we can construct an architecture equivariant under
the conformal orthogonal group.
\subsection{Graphs}
A graph is defined as $\GG(\VV,\EE)$ where $\VV=\{1,\dots, N\}$ is the
set of nodes and $\EE=\{(j,i)\}\subseteq \VV\times\VV$ is the set of
(directed) edges connecting nodes in $\VV$ (where $j$ and $i$ denote
the source and target nodes respectively). We define
$\NN_i\triangleq\{j\mid (j, i)\in\EE\}$ as the set of (in-)neighbours
of node $i$. Moreover, we associate feature node and edge embeddings,
$\vv_i\in\R^{n_v}$, $\forall i\in\VV$, and $\ee_{ji}\in\R^{n_e}$,
$\forall (j,i)\in\EE$ respectively, to each node and edge in the graph
as well as a global attribute $\uu\in\R^{n_u}$.

With a slight abuse of notation, let us define
$\GG_X(\VV,\EE)$ (often referred to simply as $\GG_X$) as a graph embedding where, in addition to the
feature node embeddings $\vv_i\in\R^{n_v}$, coordinate embeddings
$\xx_i\in\R^{n_x}$ are also associated to each node $i\in\VV$.
Moreover, let $\AA\in\VV\times\VV\times\VV$ be the set of (ordered)
triples of nodes in a graph $\GG_X$ that form an angle,  \ie,
$\AA\triangleq\{(j,i,k)\mid j,k\in\NN_i^u, j\neq k, \forall i\in
\VV\}$ with $\NN_i^u\triangleq\{j\mid (j, i)\in\EE \vee (i,j)\in\EE\}$
being the set of in- and out-neighbors
of node $i$. 

We define \emph{relative distance} and \emph{angle preserving maps} on
graph coordinate embeddings as follows.
\begin{definition}\label{def:map}
  Let $\GG_X(\VV,\EE)$ and $\GG_Y(\VV,\EE)$ be two different
  coordinate embeddings of the same graph such that $\|\xx_i-\xx_j\|^2
  = \|\yy_i-\yy_j\|^2$, $\forall (i,j)\in\EE$. Let $\xx_i^+=\psi(i,
  \GG_X)$ and $\yy_i^+=\psi(i,\GG_Y)$ for some function $\psi:\R^K\to \R^{n}$,
  $K\in \Z^+$. We say that $\psi$ is a \emph{relative distance
    preserving map} if $\|\xx_i^+-\xx_j^+\|^2 =
  \|\yy_i^+-\yy_j^+\|^2$, $\forall (i,j)\in\EE$.
\end{definition}
\begin{definition}\label{def:map}
  Let $\GG_X(\VV,\EE)$ and $\GG_Y(\VV,\EE)$ be two different
  coordinate embeddings of the same graph such that $\angle
  (\xx_j,\xx_i,\xx_k) = \angle (\yy_j,\yy_i,\yy_k)$ $\forall
  (j,i,k)\in\AA$. Let $\xx_i^+=\psi(i, \GG_X)$ and
  $\yy_i^+=\psi(i,\GG_Y)$ for some function $\psi:\R^K\to \R^{n}$,
  $K\in\Z^+$. We say that $\psi$ is a \emph{relative angle preserving
    map} if $\angle (\xx_j^+,\xx_i^+,\xx_k^+)= \angle
  (\yy_j^+,\yy_i^+,\yy_k^+)$, $\forall (j,i,k)\in\AA$.
\end{definition}
We report examples of relative distance and angle preserving maps in
Appendix~\ref{sec:maps} and their proofs.
\section{Equivariant graph networks}
We start this section by recalling the structure of a standard graph
network block. Then, we introduce two novel architectures.  The first
one is equivariant to any transformation in the node coordinates
embeddings that preserve the relative distance between neighbouring
nodes, while the second one is equivariant to transformations
preserving angles between any triple of neighbouring nodes.  In
particular, both architectures are equivariant to the Euclidean group
with the second one also equivariant to the conformal group.
Before moving further, let us stress that we will build our discussion
upon equivariance with respect to node coordinates since there is, in
general, no sense of equivariance of the network's node or edge
embeddings. However, there may be cases
in which our discussion can be extended to node or edge embeddings.

\subsection{The starting point - standard graph network block (GN)}
A general standard graph network block can be defined in terms of
edge, node and global updates as
\begin{subequations}\label{eq:gnn}
\begin{align}
  \ee_{ji}^+ &= \phi^e\big(\vv_j, \vv_i, \ee_{ji}, \uu\big),&\forall (j,i)\in\EE\\
  \vv_i^+ &= \phi^v\big(\vv_i, \rho^{e\to v}\big(\{\ee_{ji}^+\}_{j\in\NN_i}\big), \uu\big),&\forall i\in\VV\\
  \uu^+ &= \phi^u\big(\rho^{v\to u}(\{\vv_{i}^+\}_{i\in\VV}), \rho^{e\to u}(\{\ee_{ji}^+\}_{(j,i)\in\EE}), \uu\big)
\end{align}
\end{subequations}
where $\phi^e:\R^{2n_v + n_e+n_u}\to \R^{n_e^+}$, $\phi^v:\R^{n_v+
  n_e^+ + n_u}\to \R^{n_v^+}$, $\phi^u:\R^{n_v^+ + n_e^+ + n_u}\to
\R^{n_u^+}$ are update functions (usually
defined as neural networks whose parameters are to be learned) and
$\rho^{e\to v}, \rho^{e\to u}, \rho^{v\to u}$ are aggregation
functions reducing a set of elements of variable length to a single
one via an input's permutation equivariant operation like element-wise
summation or mean.

\subsection{Distance preserving graph network block (DGN)}
The first new architecture we introduce is the \emph{distance
  preserving graph network} block.  Assume that the initial node
embeddings $\vv_i$ contain no absolute coordinate or orientation
information about the initial coordinate embeddings $\xx_i$. Then, the
DGN block is defined through the following updates
 \begin{subequations}\label{eq:En}
  \begin{alignat}{3}
    &\ee_{ji}^+ &&= \phi^e\big(\ee_{ji}, \vv_i, \vv_j, \|\xx_i-\xx_j\|_2^2, \uu\big),&&\forall (j,i)\in\EE \label{eq:En_edge_update} \\
    &\vv_i^+ &&= \phi^v\big(\rho^{e\to v}\big(\{\ee_{ji}^+\}_{j\in\NN_i}\big), \vv_i, \uu\big),&&\forall i\in\VV\label{eq:En_node_update}\\
    &\xx_i^+ &&= \psi^x\left(i,\GG_X\right),&&\forall i\in\VV \label{eq:En_coord_update}\\
    &\uu^+ &&= \phi^u\big(\rho^{e\to u}(\{\ee_{ji}^+\}_{(j,i)\in\EE}),\rho^{v\to u}(\{\vv_{i}^+\}_{i\in\VV}), \rho^{x\to u}(\{\|\xx_{i}^+-\xx_j^+\|_2^2&&\}_{(j,i)\in\EE}),\uu\big)
  \end{alignat}
\end{subequations}
where $\phi^e:\R^{n_e+ 2n_v+ n_u+1}\to \R^{n_e^+}$, $\phi^v:\R^{n_e^+ + n_v+ n_u}\to \R^{n_v^+}$, $\phi^u:\R^{n_e^+ + n_v^+ + n_u+1}\to \R^{n_u^+ }$ are the update functions, $\rho^{e\to v}, \rho^{e\to u}, \rho^{v\to u}$ are aggregation functions and $\psi^x:\R^K\to \R^{n_x}$, $K\in\integers^+$ is some, possibly parametric, relative distance preserving map. 

Thanks to $\psi^x$ and the way in which node coordinates are processed to update edge and global embeddings (\ie, only through their relative distances), it can be easily seen that~\eqref{eq:En} is, by construction, equivariant to any transformation of the input coordinates that preserves their relative distances (along the edges defined by the graph structure).

\subsubsection{Euclidean group equivariance}
As a particular case, the updates~\eqref{eq:En} are also equivariant
to $E(n)$ transformations of the input coordinates. To show this, let
us start to show that the edge update~\eqref{eq:En_edge_update} is
equivariant under $E(n)$. Since $\|\xx_i-\xx_j\|_2^2$ is equivariant
under an $E(n)$ transformation, $\xx\mapsto Q\xx+\zz$, for some rotation matrix $Q\in O(n)$ and translation vector $\zz\in\R^n$, we can easily show
\begin{align*}
 \phi^e\big(\ee_{ji}, \vv_i, \vv_j, \|\xx_i-\xx_j\|_2^2, \uu\big) & \to  \phi^e\big(\ee_{ji}, \vv_i, \vv_j, \|Q\xx_i + \zz -Q \xx_j - \zz\|_2^2, \uu\big) \\
&=   \phi^e\big(\ee_{ji}, \vv_i, \vv_j, \|\xx_i - \xx_j \|_2^2, \uu\big) \,.
\end{align*}
Similarly, the coordinate
update~\eqref{eq:En_coord_update} is trivially equivariant under
$E(n)$ thanks to $\psi^x$ and the fact that only the coordinates are
affected by Euclidean transformations. Equivariance of the node
update~\eqref{eq:En_node_update} and global update follows naturally
being them functions of equivariant quantities.

\subsection{Angle preserving graph network block (AGN)}
The \emph{angle preserving graph network} block is the second novel architecture we introduce.
Given a graph $\GG_X(\VV,\EE)$, let $\aa_{jik}\in\R^{n_\alpha}$ be the
angle embedding associated to each angle $(j,i,k)\in \AA$.
Moreover, define $\AA_i$ as the set of (ordered) couples of nodes
forming an angle whose vertex is node $i$, \ie,
$\AA_i\triangleq\{(j,k)\mid (j,y,k)\in \AA, y=i\}$.
As before assume that the node attributes $\vv_i$ do not contain any
absolute coordinate information about the node coordinates
$\xx_i$. Moreover, assume that also the angle attributes $\aa_{jik}$
contain no information about the angles $\angle (\xx_j,\xx_i,\xx_k)$.

Then, the AGN block is then characterised by the following updates
\begin{subequations}\label{eq:angulaGN}
  \begin{alignat}{3}
    &\aa_{jik}^+ = \phi^\alpha(\vv_i,\vv_j,\vv_k,\aa_{jik}, \angle (\xx_j,\xx_i,\xx_k),\uu),&\forall (j,i,k)\in\AA\label{eq:ang_update}\\
    &\ee_{ji}^+ = \phi^e\big(\vv_j, \vv_i, \ee_{ji}, \uu\big),&\forall (j,i)\in\EE \\
    &\vv_i^+ = \phi^v\big(\vv_i, \rho^{e\to v}\big(\{\ee_{ji}^+\}_{j\in\NN_i}\big), \rho^{\alpha\to v}\big(\{\aa_{jik}^+\}_{(j,k)\in\AA_i}\big), \uu\big),&\forall i\in\VV\\
    &\xx_i^+ = \psi^x(i, \GG_X),&\forall i\in\VV\label{eq:coord_update_ang}\\
    &\uu^+ = \phi^u\big(\rho^{v\to u}(\{\vv_{i}^+\}_{i\in\VV}), \rho^{e\to u}(\{\ee_{ji}^+\}_{(j,i)\in\EE}), \rho^{\alpha\to u}\big(\{\aa_{jik}^+\}_{(j,i,k)\in\AA}\big), \uu\big)
  \end{alignat}
\end{subequations}
where $\phi^\alpha,\phi^e,\phi^v,\psi^x,\phi^u$ are the update functions,\footnote{The update functions in~\eqref{eq:angulaGN} live in the function spaces
  $\phi^\alpha:\R^{3n_v + n_\alpha + 1 + n_u}\to \R^{n_\alpha^+} $,
  $\phi^e:\R^{2n_v + n_e+n_u}\to \R^{n_e^+}$,
  $\phi^v:\R^{n_v+ n_e^+ + n_\alpha^+ + n_u}\to \R^{n_v^+}$,
  $\psi^x:\R^{K}\to \R^{n_x}$, for an appropriate $K\in\mathbb{Z}^+$ depending on the actual number of parameters, and 
  $\phi^u:\R^{n_v^+ + n_e^+ + n_\alpha^+ + n_u}\to \R^{n_u^+}$.}
   and $\rho^{e\to v}, \rho^{\alpha\to v},\rho^{v\to u},\rho^{e\to
u}, \rho^{\alpha\to u}$ are the aggregation functions, with $\psi^x$
  being a (possibly parametric) \emph{relative angle preserving map}.
Variants of this network that still preserve its properties can be
easily constructed by, \eg, using edge embeddings to update the angle
embeddings, using angles to update edges or not considering angle
attributes at all. Some examples are reported in
Appendix~\ref{sec:additiona_ang}.

The angular graph network is by construction equivariant to any
transformation of the coordinate embeddings that preserves the angles
created by neighbouring nodes in the graph. In particular it is
equivariant to the conformal group.
Angular information can be extremely powerful in tasks where classical
(or distance based) GNNs fail, like in graph isomorphism tests where,
\eg, a hexagon is not distinguished from two triangles.

\subsubsection{Conformal group equivariance}
As discussed in Section~\ref{sec:conformal_intro}, the conformal group
performs transformations on points whilst preserving angles between all
possible triples of coordinates (see Eq.~\eqref{eq:conf_eq}).
To show that the AGN block is equivariant to
transformations belonging to the conformal group it is sufficient to
note that given a conformal transformation $\varphi$ one has that the
angle update~\eqref{eq:ang_update} is equivariant to it since
\begin{align*}
  \phi^\alpha(\vv_i,\vv_j,\vv_k,\aa_{jik}, \angle (\xx_j,\xx_i,\xx_k),\uu) &\to \phi^\alpha(\vv_i,\vv_j,\vv_k,\aa_{jik}, \angle (\varphi(\xx_j),\varphi(\xx_i),\varphi(\xx_k)),\uu)\\
  &=\phi^\alpha(\vv_i,\vv_j,\vv_k,\aa_{jik}, \angle (\xx_j,\xx_i,\xx_k),\uu) \,.
\end{align*}
The coordinate update~\eqref{eq:coord_update_ang} is additionally
equivariant to the conformal group thanks to the assumption on
$\psi^x$ and the fact that only coordinates are affected by a
conformal transformation. Equivariance of the other updates is
trivially satisfied by construction.
We stress that equivariance to the conformal group includes by
definition equivariance to the Euclidean group as we are only
concerned with transformations on $\R^n$. The orthogonal rotations and
translations of the Euclidean group are therefore a subset of possible
conformal transformations on $\R^n$ as they are angle-conserving
transformations.

\section{Discussion}
  In the previous section we introduced two novel graph
  architectures. To construct them we have mainly built upon the
  actions on a coordinate system of two transformations that preserves
  distances and angles between neighbouring nodes in a graph. These transformations include as particular cases the Euclidean
  and conformal groups. Equivariance under $E(n)$ means equivariance
  to orthogonal rotations and translations while by considering the
  full conformal group, we can further generalise our neural network
  architecture. Conformal $n$-dimensional transformations consist of
  the groups containing translations, dilations, rotations and
  inversions with respect to an $n-1$ sphere. A conformal
  transformation is therefore a powerful tool for mapping data points
  onto each other, and hence, building a neural network architecture
  equivariant to the conformal group enables the architecture to be
  equivariant to a wide selection of interesting subgroups.
  By introducing distance and angle preserving transformations that
  take into account the structure of the graph, equivariance under
  more general transformations have been achieved. A notable example
  is that the DGN architecture is equivariant to any transformation of
  a Hoberman sphere, while both the DGN and AGN can be equivariant to
  transformations affecting only subsets of their nodes, as shown
  below.

In summary, by taking advantage of the powerful tools of group theory,
and performing operations on a vector space which are equivariant under group
transformations, we can build a neural network architecture which can
take advantage of these group properties. Such an architecture will be
able to deal with rotated, translated, dilated (or more generally
transformed) data more efficiently than a standard graph network which
does not have the above group properties built into it.

\begin{figure}
  \centering
  \begin{subfigure}{0.4\textwidth}
    \centering
    \includegraphics[width=0.55\textwidth]{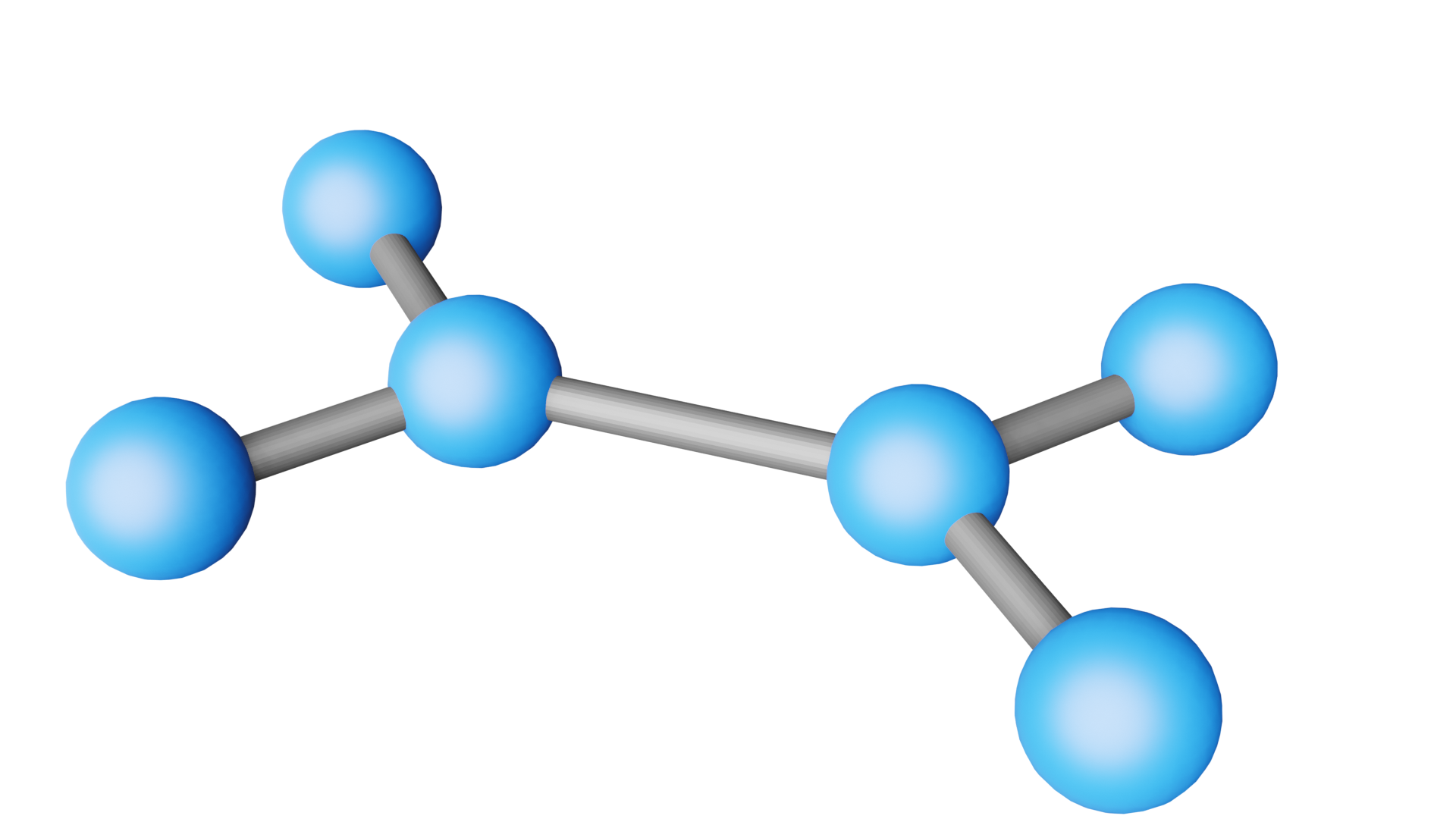}
  \end{subfigure}
  \begin{subfigure}{0.4\textwidth}
    \centering
    \includegraphics[width=0.55\textwidth]{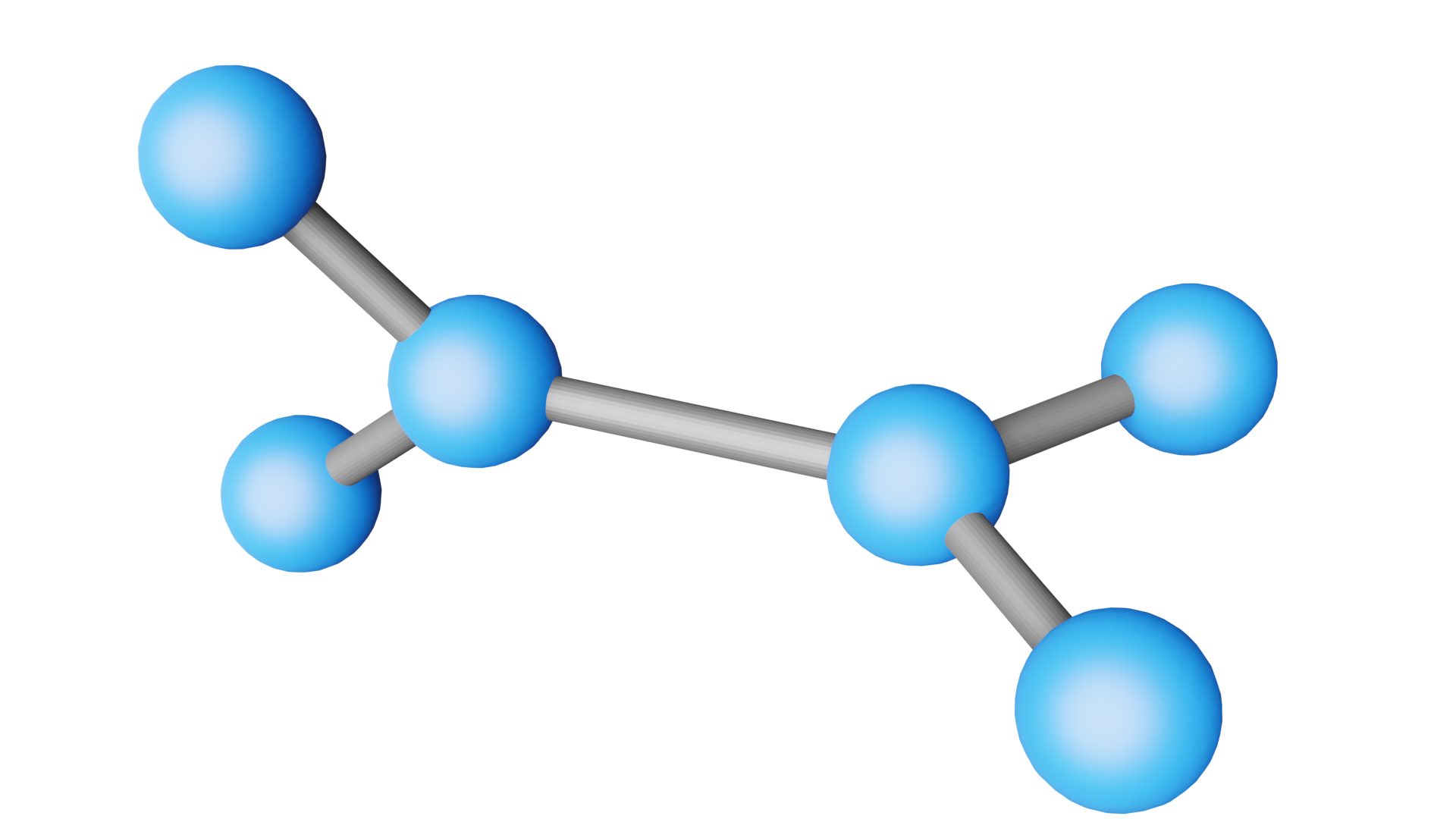}
  \end{subfigure}
  \caption{Example of local transformations. Each one of the two 3-nodes subgraphs composing the graph can be rotated without affecting distances or angles between neighbouring nodes.}
  \label{fig:conformation}
\end{figure}
\subsection{Beyond global transformations}
So far, we have talked about global transformations in the coordinate space. In general, a global symmetry $\varphi_g$, as defined in Section~\ref{sec:groups} acts on a function $\psi(\xx)$ as
\begin{equation}
\psi(\xx) \to \varphi_g \psi(\xx) \,.
\end{equation}
A \emph{local} group symmetry $\varphi_l (\xx) $ is instead defined as
\begin{equation}
\psi(\xx) \to \varphi_l (\xx) \psi(\xx) \,,
\end{equation}
where now the group action can differ across all points in the space (see \eg~\citep{Peskin:1995ev} for an
in-depth discussion).

The architectures we present are also able to deal with some local
symmetries. Namely, the DGN and AGN preserve local distances and
angles respectively \emph{where a suitable subgraph must be defined}.
As an example consider a transformation that rotates only some nodes
of a graph; all nodes which are in the neighbourhoods of the nodes
will, in the angular architecture, have their angles preserved, and in
the distance-preserving architecture, the distances between nodes will
be preserved.
A pertinent use-case for such a symmetry is that of molecular
conformations, where the spatial arrangement of the atoms can be
interconverted by rotations about formally single bonds, as shown in
Figure~\ref{fig:conformation}). The AGN would be equivariant to such a
transformation at the level of the two distinct subgraphs separated by
the bond in question, thus enabling one to learn
conformation-invariant properties.

\subsection{Comparing and extending the two architectures}\label{sec:comparison}
\begin{figure}
  \centering
  \includegraphics[width=0.63\textwidth]{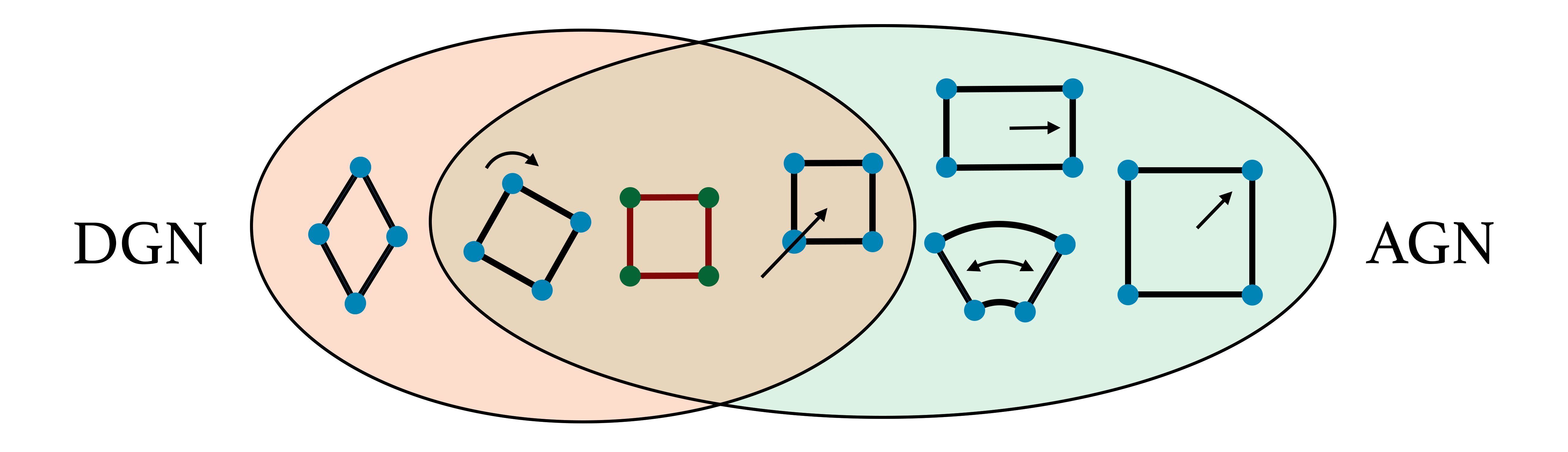}
  \caption{Equivariance of the two networks. The green/red square denotes the base graph and the blue/black ones alternative coordinate embeddings, equivariant for the DGN (red) or AGN (green).}
  \label{fig:comp}
\end{figure}

As mentioned, the two architectures we presented are partially
overlapping in terms of equivariance features. Both of them are
equivariant to the $E(n)$ group but the first one can deal with
non-orthogonal transformations that preserve the distance between
neighbouring nodes (such as with the Hoberman sphere) and the angular
one is equivariant to the conformal group which allows to
non-orthogonal transformations that preserve angles but not distances
(see Figure~\ref{fig:comp}).

The distance equivariant graph network can be equipped with a sense of
scale equivariance through an input scaling layer that makes it
invariant to the conformal orthogonal group. Let $\gamma = \alpha /
\max_{(i,j)\in\EE} \|\xx_i-\xx_j\|$ for some $\alpha\in \R$, where we
identify $\gamma$ as satisfying the dilation condition required under
the conformal orthogonal group. Then, conformal orthogonal invariance
can simply be obtained by computing scale-normalised coordinates
$\tilde{\xx}_i=\gamma \xx_i$ and using them as the coordinates
in~\eqref{eq:En}.

Introducing coordinate scaling therefore enables the distance
equivariant graph network to be scale equivariant, under the condition
that all transformations remain orthogonal. With the AGN, we can relax
this condition and be equivariant under the full conformal group. This
larger group equivariance enables us to work with very powerful
transformations on data, which can also be dangerous. For example, if
the scale is an important property of the dataset, the AGN will not
recognise it and so will learn incorrect information about the
data. Similarly, one must also not use the DGN blindly and
irrespective of the properties of the dataset; as the architecture is
equivariant to transformations that preserve distance between
neighbouring nodes but not angles it is able to, for example, map a
square onto a line. The two architectures we presented can be also
combined together at the price of losing some features, while
generalising to both distance and angle preserving transformations.

\section{Related work}
The study and formulation of group equivariant neural networks have flourished in the last years and they have proven to be
extremely powerful in many tasks.
The first example is probably that of convolutional neural
networks~\citep{lecun1990handwritten}, which are translation
equivariant, thanks to the convolution operation, and have led to
breakthroughs in most vision tasks. CNNs have been generalised to
exploit larger groups of symmetries. G-CNNs are proposed
in~\citep{cohen2016group} to deal with more general symmetries,
including rotations and translations, and extended
in~\citep{Bekkers2020B-Spline,finzi2020generalizing} to deal with Lie
groups. Equivariance to arbitrary symmetry groups can be achieved via
self-attention mechanisms in~\citep{romero2021group}.  Continuous
convolutions are used in SchNet~\citep{schutt2017quantum} to achieve
$E(n)$ invariance, and $SE(3)$ equivariance is achieved
in~\cite{thomas2018tensor} via the use of spherical harmonics.  The
drawback of many of these methods is their limited
applicability due to computational complexity.

There has been some work on constructing general MLPs that are
equivariant to different groups; a layer equivariant to general matrix
groups is presented in~\cite{finzi2021practical}, whilst equivariance
to the Lorentz group for physics applications has recently been
explored~\citep{bogatskiy2020lorentz}. Further applications
include~\citep{mattheakis2019physical}, where physical symmetries are
embedded in neural networks via embedding physical constraints in the
structure of the network, and in~\citep{barenboim2021symmetry} where
the ability of neural networks to discover and learn symmetries in
data is explored.

Graph neural networks are, by construction, permutation
equivariant~\citep{scarselli2008graph,battaglia2018relational}. Recently,
there has been a lot of work on building equivariance to other
interesting groups in GNNs. There has been particular interest in the
Euclidean group with results for the subgroups $SE(3)$ and $E(3)$
obtained in~\citep{1802-08219,NEURIPS2020_15231a7c,
  kohler2020equivariant, finzi2020generalizing,
  batzner2021se3equivariant, Yang_2020_CVPR}.
 Extending these architectures, an $E(n)$ equivariant message passing convolutional graph layer is proposed in~\citep{satorras2021en}, leading to state of the art results in a variety of tasks, including $n$-body particle systems simulations and molecular property predictions. A similar idea has been also proposed in~\citep{slade2021permutation}, where a general $E(n)$ equivariant GNN is presented, and in~\citep{horie2021isometric}, in which equivariance to isometric transformations is considered. These $E(n)$ equivariant networks can be seen as special instances of our distance equivariant architecture, where particular aggregation functions are applied and a particular choice is made for the distance preserving map $\psi^x$ (see Appendix~\ref{sec:other}). In~\citep{satorras2021en}, for example, the map $\psi^x$ is chosen so that relative distances between all possible pairs of node in the graph are preserved, thus allowing only for rigid transformations in $E(n)$.

  Angular information is used in~\citep{smith2017ani} but no attention is explicitly paid to equivariance. DimeNET is proposed in~\citep{klicpera2020directional}, where both distance and angle embeddings computed through embeddings in novel orthogonal basis functions are used in a message passing graph network to achieve equivariance. State of the art results are obtained on molecular property and dynamics prediction datasets. We note, however, that while DimeNET produces state of the art results, it is restricted to atomic data, due to the requirement of Gaussian radial and Bessel functions as the basis representations.
  Additionally, angular embeddings computed in the isometric invariant layer in~\citep{horie2021isometric} corresponds to the extraction of both relative distances and angles of each pair of vertices.
  Again, these approaches are special instances of our networks where particular aggregation and learnable functions or hand engineered distance and/or angle embeddings are used (see Appendix~\ref{sec:other}). In this sense, our network is unconstrained and can learn what it needs. Moreover, thanks to its general form it can achieve equivariance to the conformal group (and more) which, to the best of our knowledge, does not hold for known architectures.
  
  To summarise, most of the existent equivariant graph architecture are at most equivariant to the Euclidean group (represented as the intersection set in Figure~\ref{fig:comp}), while the architectures proposed here are, in their general form, equivariant to more much general transformations.

\section{Experimental results}
In this section we give an insight into the performance of the architectures we proposed and on their limitations. 
First, we demonstrate that perfect generalisation to unseen data can be achieved on datasets with a large number of symmetries, possibly accompanied by a faster convergence rate (as shown for QM9 in Appendix~\ref{sec:qm9_results}).
Then, we show that, when used on datasets with few symmetries, using our architectures can be counterproductive.
In general, when the right symmetries are present, equivariant architectures experience both increased accuracy and convergence speed, as demonstrated by state of the art results obtained on a number of tasks by architectures that can be obtained as particular instances of the ones proposed here. Examples include $n$-body system dynamics prediction, molecular dynamics and molecular property prediction tasks~\citep{satorras2021en,klicpera2020directional,horie2021isometric}. 

\begin{table}
  \centering
  \tiny
  \begin{tabular}{l|c|c|c|c|c|c|c|}
    & & & \multicolumn{5}{|c|}{test accuracy}  \\\cline{4-8}
    block & $\rho$ & train acc & Orthogonal  & Orthogonal + dilation  & \begin{tabular}{c}Non-orthogonal\\($\mu=0.5$) \end{tabular} & \begin{tabular}{c}Non-orthogonal\\($\mu=1.5$) \end{tabular} & \begin{tabular}{c}Non-orthogonal\\($\mu=3.0$) \end{tabular}\\
    \hline\hline
    AGN & mean      & $1$   & $\bm{1}$        & $\bm{1}$        & $\bm{1}$       & $\bm{1}$       & $\bm{0.96\pm0.04}$ \\
    AGN & sum       & $1$   & $\bm{1}$        & $\bm{1}$        & $\bm{1}$       & $\bm{1}$       & $\bm{1}$            \\
    SDGN & mean     & $0.2$ & $0.2$           & $0.2$           & $0.2$          & $0.2$          & $0.2$ \\
    SDGN & sum      & $1$   & $\bm{1}$        & $\bm{1}$        & $\bm{1}$       & $0.91\pm 0.07$ & $0.83\pm 0.11$\\
    DGN & mean      & $1$   & $\bm{1}$        & $0.28\pm 0.02$  & $0.20\pm 0.01$ & $0.28\pm 0.01$ & $0.30\pm 0.01$ \\
    DGN$^*$ & sum       & $1$   & $\bm{1}$        & $0.34\pm 0.05$  & $0.28\pm 0.12$ & $0.36\pm 0.07$ & $0.35\pm 0.04$ \\
    GN & mean       & $1$   & $0.26\pm 0.03$  & $0.25\pm 0.03$  & $0.27\pm 0.04$ & $0.25\pm 0.03$ & $0.24\pm 0.04$ \\
    GN & sum        & $1$   & $0.40\pm 0.05$  & $0.36\pm 0.04$  & $0.32\pm 0.05$ & $0.38\pm 0.07$ & $0.35\pm 0.05$ \\
    \hline
  \end{tabular}
  \caption{Polytopes classification: training and test accuracy (mean $\pm$ standard deviation over $10$ runs) for $n=3$, for different transformations in the test set. $^*$ equivalent to~\citep{satorras2021en}.}
  \label{tab:polytope_orth}
\end{table}

\subsection{Polytopes classification} \label{sec:shapes_results}
We consider a $n$-dimensional polytopes classification problems for $n=3,4,5$. The datasets are composed of graph representations of polytopes and the number of classes varies with $n$. For $n=3$, simplexes, hypercubes, orthoplexes, dodecahedra and icosahedra are considered. Details and results for other values of $n$ are reported in Appendix~\ref{sec:poly}.
The training dataset consists of a single graph per regular polytope (specified in terms of node coordinates and list of edges) while the test set is composed of randomly transformed versions of those in the training set, whose coordinates are obtained as $\tilde{\xx}_i = \gamma A\xx_i + q$ for some $\gamma\in\R$, $A\in\R^{n\times n}$ and $q\in\R^n$.
We compare graph networks built with AGN, DGN and standard blocks. The DGN block is also combined with a scaling layer (SDGN) as in Section~\ref{sec:comparison} to obtain equivariance to dilations. Mean and sum aggregation functions $\rho$ are considered. Results are reported in Table~\ref{tab:polytope_orth} for $\psi^x$ chosen as the identity function, \ie, $\xx_i^+=\xx_i$ (a more complex one is also considered in Appendix~\ref{sec:poly}).

We note that all models reach a perfect accuracy on the training set, except the SDGN with mean aggregation function, which is not able to distinguish any polytopes. If one changes the coordinate update function $\psi^x$ the accuracy reaches $0.8$ and only simplexes and orthoplexes are confounded. Adding (non all identical) node attributes to the graphs in addition to node coordinates enables all polytopes to be discerned. This behaviour is caused by the combination of the mean aggregation function and the scaling layer which makes the information propagated through the network be the same for all polytopes.
When only orthogonal transformations are performed in the test set (\ie, $A^\top A=I$, $\gamma=1$) AGN, DGN and SDGN (with sum aggregation) perfectly generalise to unseen data. However, as expected, when also adding  dilations (\ie, $A^\top A=\gamma I$, $\gamma\in\R$), only AGN and SDGN (with sum aggregation) can generalise well. Finally, when adding random non-orthogonal transformations (\ie, $\mu = \text{E}[\|A^\top A - I\|_F]>0$, $\gamma\in\R$), something possibly unexpected can be seen to happen. The AGN (and partially also the SDGN) perfectly generalises to unseen polytopes whose angles have not been preserved (due to the non-orthogonality of $A$). This is probably due to the network having learnt to distinguish polytopes based on the sum (or mean) of their angles which is preserved under non-orthogonal transformations due to the Gram–Euler theorem. While this behaviour turns out to be useful in the task at hand, one can easily envisage cases in which it may cause issues.

\subsection{Benchmark datasets}
Now we consider benchmark datasets from~\citep{dwivedi2020benchmarking} for which node coordinates are provided. In particular, we consider MNIST, CIFAR10 as graph classification problems and TSP as an edge classification one. While, when treated as images, MNIST and CIFAR10 are usually examples of datasets containing symmetries, in their graph representation such symmetries are mostly lost. Furthermore, in TSP there are no apparent symmetries in the dataset. In these cases, as one may expect, using an equivariant network would not produce any benefits. Rather, as we show, negative effects can appear.

We trained our models and a standard GNN on these datasets. All models have roughly the same number of parameters and the results are reported in Table~\ref{tab:benchmark}. As  can be seen, treating node coordinates as if they can posses some symmetric between among different samples turns out to be counterproductive both in terms of convergence speed and accuracy.
In particular we see that while the standard GN has the best performance, the DGN is slightly better than the AGN. This is probably due to the fact that, while some non-orthogonal transformations mostly preserving local distances may be present in the dataset, angle preserving ones are much more rare. On TSP in particular, using angle related properties results in a significant drop in performance.

\begin{table}
  \centering
  \tiny
  \begin{tabular}{l|c|c|c|c|c|c|}
           & \multicolumn{2}{|c|}{MNIST}                   & \multicolumn{2}{|c|}{CIFAR10}                 & \multicolumn{2}{|c|}{TSP}                    \\\cline{2-7}
    block  & train acc             & test acc              & train acc             & test acc              & train F1 (positive)   & test F1 (positive)                \\
    \hline\hline
    AGN    & $0.943\pm 0.004$      & $0.932\pm 0.004$      & $0.644\pm 0.004$      & $0.590\pm 0.004$      & $0.632\pm 0.011$      & $0.681\pm 0.020$          \\
    DGN    & $0.957\pm 0.001$      & $0.945\pm 0.006$      & $0.657\pm 0.004$      & $0.592\pm 0.002$      & $0.748\pm 0.013$      & $0.771\pm 0.014$          \\
    GN     & $\bm{0.982\pm 0.001}$ & $\bm{0.977\pm 0.002}$ & $\bm{0.719\pm 0.007}$ & $\bm{0.657\pm 0.001}$ & $\bm{0.772\pm 0.041}$ & $\bm{0.793\pm 0.033}$     \\
    \hline
  \end{tabular}
  \caption{Benchmark datasets. Train and test accuracy are reported for MNIST and CIFAR10. For TSP, due to the high class unbalance, the train and test F1 score for the positive class is reported.}
  \label{tab:benchmark}
\end{table}

 \subsubsection{Computational complexity}
All the experiments have been run on an NVIDIA V100 GPU. The average time required for each training step (\ie, forward pass, back-propagation and optimiser step for a batch of data) is reported in Table~\ref{tab:time}, normalised with respect to the batch-size.
Using the scaling layer for the DGN drastically increase the computational complexity.
Moreover, it can be seen that while the AGN and DGN have similar complexities for dataset in which the number of edges (and hence angles) is relatively low, when graphs are (nearly) complete (like in TSP) or highly connected, the AGN has an overhead with respect to the DGN (due to the need of updated and computing angles in addition to edges).

\begin{table}
  \centering
  \tiny
  \begin{tabular}{l|c|c|c|c|c|c|c|}
        & \multicolumn{7}{|c|}{Time per training step (ms)}  \\\cline{2-8}
        & polytopes (n=3) & polytopes (n=4) & polytopes (n=5) & MNIST & CIFAR10 & TSP & QM9\\
  batch-size & 5 & 6 & 3 & 128 & 128 & 8 & 512\\
  \hline\hline
  AGN   & 3.8       & 3.8         & 6.3        & 1.17 & 1.64 & 35  & 0.23\\
  SDGN  & 5         & 14.5        & 8          & 7.81 & 12.5   & - & -\\
  DGN   & 3.8       & 3.7         & 6.3        & 0.78 & 1.09  & 16 & 0.19\\
  GN    & 3         & 3           & 5          & 0.77 & 1.02  & 14  & 0.16\\
  \hline  
  \end{tabular}
  \caption{Time per training step in the various experiments, normalised with respect to the batch-size.}
  \label{tab:time}
\end{table}

\section{Conclusion}
In this paper we have presented novel deep learning architectures which are equivariant to distance and angle preserving transformations in graph coordinate embeddings. In particular, we have shown equivariance to the $E(n)$, $\CO(\R^n, \mathcal{Q})$ and $\text{Conf}(\R^{n,0})$ groups in addition to permutation equivariance.
Equivariance to local symmetries can also be achieved when the different transformations are applied to suitable subgraphs.
We have applied our models to a synthetic dataset composed of $n$-dimensional regular polytopes as well as to several benchmark datasets.
We have shown that the architectures we propose are significantly more accurate and data efficient than a standard graph network on datasets where there are large numbers of symmetries in the data. We also explicitly show examples where our architecture produces unexpected results or would not be applicable, due to a lack of symmetry in the data.

\bibliographystyle{plainnat}
\bibliography{biblio.bib}

\appendix

\section{$E(n)$ equivariance of $\|\xx_i - \xx_j\|_2^2$}\label{sec:En_eq}
To show that $\|\xx_i - \xx_j\|_2^2$ is equivariant under $E(n)$, it is sufficient to see that, under translation, one has
  \begin{equation*}
    \xx \to \xx^+= \xx + \zz, \quad \zz \in \E^n \,,
  \end{equation*}
while under rotation
  \begin{equation*}
    \xx \to \xx^+= Q\xx, \quad Q \in O(n) \,.
  \end{equation*}
Now, by using the above relations and the fact that $E(n)$ is the semidirect product of $T(n)$ and $O(n)$, it is easy to show that $\|\xx_i-\xx_j\|_2^2$ is equivariant under $E(n)$ since
\begin{subequations}\label{eq:En_eq}
\begin{align}
  \|\xx_i - \xx_j\|_2^2\to \|\xx_i^+ - \xx_j^+\|_2^2 &= (Q(\xx_i + \zz) - Q(\xx_j + \zz) )^\top(Q(\xx_i + \zz) - Q(\xx_j + \zz) )\\
    &= (Q\xx_i - Q\xx_j  )^\top(Q\xx_i  - Q\xx_j ) \\
    &= (\xx_i - \xx_j  )^\top Q^\top Q (\xx_i  - \xx_j ) \\
    &= (\xx_i - \xx_j  )^\top (\xx_i  - \xx_j ) \,,
\end{align}
where we used the fact that $Q^\top Q=I \, \forall Q \in O(n)$.
\end{subequations}

\section{Equivariance under the conformal orthogonal group}\label{sec:CO}
In this section we will derive the result from Section 2.1.2 in the main text, that dilations and orthogonal rotations are transformations under $\CO(\R^n, \mathcal{Q})$. The group is defined for a vector space $V$ with a quadratic form $\mathcal{Q}$. The group contains the linear transformations $\varphi : \mathcal{T} \to V$. The group action is defined as 
\begin{equation} \label{eq:conformal_group}
\mathcal{Q}(\mathcal{T}x) = \gamma^2 \mathcal{Q}(x) \,,
\end{equation}
where $\mathcal{T}$ is the set of linear transformations that we need to define and $\gamma$ is a scalar.
For our purposes, we can consider a positive-definite quadratic form on $\R^n$; as we restrict our discussion to Euclidean space then the relevant quadratic form is
\begin{equation} \label{eq:euclidean_distance}
\mathcal{Q} = \sum_i x_i^2 \,,
\end{equation}
and so inserting~\eqref{eq:euclidean_distance} into~\eqref{eq:conformal_group} we have the condition
\begin{equation*}
\sum_i (\mathcal{T}  x_i)^2 = \gamma^2 \sum_i x_i^2 \,.
\end{equation*}
 If we consider, as above, an orthogonal transformation $Q$, and in addition, a dilation $x_i \to \gamma x_i$, then it is simple to show that, with $\mathcal{T} = \gamma Q$
\begin{align*}
\sum_i (\gamma Q  x_i)^2 &= \gamma^2 Q^\top Q \sum_i x_i^2 = \gamma^2  \sum_i x_i^2 \,.
\end{align*}

\section{Relative distance and angle preserving maps}\label{sec:maps}
Many different relative distance and/or angle preserving maps to be used in the coordinate update can be obtained in different ways, eventually requiring further assumptions on the data.

The simplest one is the identity function
\begin{equation}\label{eq:map_I}
  \xx_i^+ = \psi(i,\GG_X) = \xx_i
\end{equation}
which keeps the coordinates unchanged during the update step. In this case both relative distances and angles are trivially preserved. Updating all the coordinates in the same way also trivially preserves both relative distances and angles. Examples include
\begin{align*}
  \xx_i^+ &= a \xx_i, &&a\in\R,\\
  \xx_i^+ &= \xx_i + \mathbf{a}, &&\mathbf{a}\in\R^{n_x}\\
  \xx_i^+ &= Q \xx_i + \mathbf{a}, &&Q\in\R^{n_x\times n_x}, Q^\top Q=I,  \mathbf{a}\in\R^{n_x}
\end{align*}
where $a, \mathbf{a}, Q$ can be learnable parameters or parametric functions of other network parameters, \eg, $\mathbf{a}=\phi^x(\uu)$. The matrix $Q$ can also be non orthogonal, provided that the obtained transformation preserves distances or angles.

Devising more complex forms for $\psi$ for arbitrary coordinate embeddings $X$, $Y$ satisfying the relative distance or angle preserving property, is quite tricky and further assumptions are in general required.
To see this, consider a relative distance preserving transformation. Any transformation $X\to Y$, such that $\|\xx_i-\xx_j\|^2 = \|\yy_i-\yy_j\|^2$, can be obtained as $\xx_i=\gamma_iA_i\yy_i + \qq_i$, where $\gamma_i$, $A_i$ and $\qq_i$, $i\in\VV$, are solutions to the system
\begin{equation*}
    \|\gamma_i A_i\yy_i + \qq_i - \gamma_j A_j\yy_j + \qq_j\|^2 = \|\yy_i-\yy_j\|^2, \quad (i,j)\in\EE.
\end{equation*} 
with $\gamma_i\in\R$, $A_i^\top A_i=I$, $\qq_i\in\R^{n_n}$, $\forall i\in\VV$. In this general case, defining a (non trivial) map $\psi$ such that, after its application, one has $\|\xx_i^+-\xx_j^+\|^2 = \|\yy_i^+-\yy_j^+\|^2 \forall (i,j)\in\EE$ is hard without any assumption on, at least, the topology of the graph $\GG_X$ (or, equivalently $\GG_Y$). A similar reasoning can be applied also to relative angle preserving maps.

If we restrict ourselves to the case in which $\gamma_i=\gamma$, $A_i=A$, $\qq_i=\qq$, $\forall i\in\VV$, this results in $X$ being a Conformal orthogonal transformation of $Y$. In this case, a possible map $\psi$ is defined by
\begin{equation}
  \xx_i^+ = \xx_i + \sum_{j \in \NN_i} a_{ji}(\xx_j-\xx_i).\label{eq:eq_map}
\end{equation}
with $a_{ji}$ possibly being a parametric function of node/edge/angle/global attributes, \eg, $a_{ji}=\phi^x\left(\ee_{ji}^+,\vv_j^+,\vv_i^+, \uu\right)$.
Now we show that~\eqref{eq:eq_map} is both relative distance and angle preserving. For the sake of notation, we assume $\gamma=1$, however the same arguments can be applied when $\gamma\neq 1$.

\paragraph{Relative distance preservation of~\eqref{eq:eq_map}}
To show that~\eqref{eq:eq_map} satisfies the definition of a distance-preserving map it is sufficient to show that, since $\xx_i=A\yy_i + q$, one has
\begin{align}
  \xx_i^+ &=\xx_i + \sum_{j \in \NN_i} a_{ij}(\xx_j-\xx_i) \nonumber\\
  &= A\yy_i+\qq + \sum_{j \in \NN_i} a_{ij}(A\xx_j+\qq-A\xx_i-\qq)\nonumber\\
  &=A\yy_i+\qq + \sum_{j \in \NN_i} a_{ij}(A\xx_j-A\xx_i)\nonumber\\
  &=A\left[\yy_i + \sum_{j \in \NN_i} a_{ij}(\xx_j-\xx_i)\right] +\qq\nonumber\\
  &=A\yy_i^+ +\qq\label{eq:En_tran}
\end{align}
which implies
\begin{align*}
  \|\xx_i^+ - \xx_j^+\|^2 &= \|A\yy_i^+ + \qq - A\yy_j^+ - \qq\|^2\\
  &= \|A(\yy_i^+  - \yy_j^+ )\|^2\\
  &=(\yy_i^+  - \yy_j^+ )^\top A^\top A (\yy_i^+  - \yy_j^+ )\\
  &= \|\yy_i^+  - \yy_j^+\|^2
\end{align*}
where in the last line we used the fact that $A^\top A=I$.

\paragraph{Relative angle preservation of~\eqref{eq:eq_map}}
While~\eqref{eq:En_tran} implies that angles are preserved since $\xx_i$ is an $E(n)$ transformation of $\yy_i$, one can show this explicitly by recalling that the angle between two vectors $\xx_j-\xx_i$ and $\xx_k-\xx_i$ can be computed as
  \begin{align*}
    \text{cos}\angle(\xx_j,\xx_i,\xx_k) &= \frac{(\xx_j-\xx_i)^\top (\xx_k-\xx_i)}{\|\xx_j-\xx_i\|\|\xx_k-\xx_i\|}
  \end{align*}
  Then, one has 
  \begin{align*}
    \frac{(\xx_j^+-\xx_i^+)^\top (\xx_k^+-\xx_i^+)}{\|\xx_j^+-\xx_i^+\|\|\xx_k^+-\xx_i^+\|} &=  \frac{(A\yy_j^+ +\qq-A\yy_i^+ -\qq)^\top (A\yy_k^+ +\qq-A\yy_i^+ -\qq)}{\|A\yy_j^+ +\qq-A\yy_i^+ -\qq\|\|A\yy_k^+ +\qq-A\yy_i^+ -\qq\|}\\
    &=  \frac{(A\yy_j^+ -A\yy_i^+ )^\top (A\yy_k^+ -A\yy_i^+ )}{\|A\yy_j^+ -A\yy_i^+ \|\|A\yy_k^+ -A\yy_i^+ \|}\\
    &=  \frac{(\yy_j^+ -\yy_i^+ )^\top A^\top A (\yy_k^+ -\yy_i^+ )}{\sqrt{(\yy_j^+ -\yy_i^+ )^\top A^\top A(\yy_j^+ -\yy_i^+ )} \sqrt{(\yy_k^+ -\yy_i^+ )^\top A^\top A(\yy_k^+ -\yy_i^+ )}}\\
    & =  \frac{(\yy_j^+-\yy_i^+)^\top (\yy_k^+-\yy_i^+)}{\|\yy_j^+-\yy_i^+\|\|\yy_k^+-\yy_i^+\|}
  \end{align*}
  where in the last line we used the fact that $A^\top A=I$.

\paragraph{Local symmetry transformation of~\eqref{eq:eq_map}}
Suppose we have a local symmetry transformation, $A(\tilde{\xx})$ which only acts on a subgraph $\tilde{\GG} \in \GG$, such that $\vv = (\tilde{\vv}_1, \tilde{\vv}_2, \dots \tilde{\vv}_n, \vv_{n+1}, \dots \vv_m)$, and similarly for the edge, coordinate and angle features. The action of $A(\tilde\xx)$ is then
\begin{align*}
A(\tilde{\xx})\xx &= \left( A(\tilde{\xx})\tilde{\xx}_1, A(\tilde{\xx})\tilde{\xx}_2, \dots A(\tilde{\xx})\tilde{\xx}_n, A(\tilde{\xx})\xx_{n+1}, \dots A(\tilde{\xx})\xx_m  \right) \\
&= \left( A(\tilde{\xx})\tilde{\xx}_1, A(\tilde{\xx})\tilde{\xx}_2, \dots A(\tilde{\xx})\tilde{\xx}_n, \xx_{n+1}, \dots \xx_m  \right) \,.
\end{align*}

By defining the 2 subgraphs as containing the coordinate features which are and are not affected by the symmetry transformation, we can therefore write a coordinate update~\eqref{eq:eq_map} for both subgraphs; Eq.~\eqref{eq:eq_map} for the $\xx_i$ and, for the $\tilde{\xx}_i$,
\begin{equation*}
  \tilde{\xx}_i^+ =\tilde{\xx}_i + \sum_{j \in \NN_i} a_{ji}(\tilde{\xx}_j-\tilde{\xx}_i).
\end{equation*}
We have shown above that this coordinate update preserves distances and angles for global group transformations; by defining the 2 subgraphs as above, we can promote the local symmetry transformation $A(\tilde{\xx})$ to global transformations on subgraphs. Whilst here we have shown this for only 2 subgraphs, one can generalise the argument to any number of local transformations so long as the subgraphs are defined as above. As we have to subdivide the graph into subgraphs, we note that these local transformations cannot be defined arbitrarily as there must be a sense of a neighbourhood of nodes within each subgraph. A local transformation which affects unrelated nodes identically (which is a valid class of local symmetry) is not valid for this reason.

\section{Additional formulations}
  \subsection{Alternative formulations for the angle preserving graph network}\label{sec:additiona_ang}
  A number of variations can be proposed for the angle preserving graph network:
  \begin{itemize}
    \item Edge attributes can be used in angle updates
    \begin{align*}
      \aa_{jik}^+ &= \phi^\alpha(\vv_i,\vv_j,\vv_k,\ee_{ij},\ee_{ik}, \aa_{jik}, \angle (\xx_j,\xx_i,\xx_k),\uu),&\forall (j,i,k)\in\AA.
    \end{align*}
    \item Relative distances can be used in the angle updates
    \begin{align*}
      \aa_{jik}^+ = \phi^{\alpha} (\dots,\|\xx_i-\xx_j\|_2^2,\|\xx_i-\xx_k\|_2^2,\angle (\xx_j,\xx_i,\xx_k), \uu),\,\,\,\, &&\forall (j,i,k)\in\AA.
    \end{align*} 
    \item Angle attributes can be used in edge updates
    \begin{align*}
      \ee_{ij}^+ &= \phi^e\big(\ee_{ij}, \vv_i, \vv_j, \rho^{\alpha\to e}\big(\{\aa_{ijk}^+\}_{k\in\AA_{ij}}\big), \rho^{\alpha\to e}\big(\{\aa_{jik}^+\}_{k\in\AA_{ji}}\big), \uu\big),&\forall (i,j)\in\EE
    \end{align*}
    where $\AA_{ij}=\{k \mid (y,z,k)\in \AA, y=i, z=j\}$ is the set of angles whose first ray is defined by $(i,j)$.
    \item Angle embeddings can be ignored and node attributes can be updated with the angles themselves
    \begin{subequations}
      \begin{alignat*}{3}
        &\ee_{ji}^+ = \phi^e\big(\vv_j, \vv_i, \ee_{ji}, \uu\big),&\forall (j,i)\in\EE \\
        &\vv_i^+ = \phi^v\big(\vv_i, \rho^{e\to v}\big(\{\ee_{ji}^+\}_{j\in\NN_i}\big), \rho^{\alpha\to v}\big(\{\angle (\xx_j,\xx_i,\xx_k)\}_{(j,k)\in\AA_i}\big), \uu\big),&\forall i\in\VV\\
        &\xx_i^+ = \psi^x(i, \GG_X),&\forall i\in\VV\\
        &\uu^+ = \phi^u\big(\rho^{v\to u}(\{\vv_{i}^+\}_{i\in\VV}), \rho^{e\to u}(\{\ee_{ji}^+\}_{(j,i)\in\EE}), \uu\big).
      \end{alignat*}
    \end{subequations}
  \end{itemize}

  \subsection{Combined architecture}
  The DGN and AGN architectures can be combined in a single architecture. An example is
  \begin{subequations}
    \begin{alignat*}{3}
      &\aa_{jik}^+ = \phi^\alpha(\vv_i,\vv_j,\vv_k,\aa_{jik}, \angle (\xx_j,\xx_i,\xx_k),\uu),&\forall (j,i,k)\in\AA\\
      &\ee_{ji}^+ = \phi^e\big(\vv_j, \vv_i, \ee_{ji}, \|\xx_i-\xx_j\|_2^2, \uu\big),&\forall (j,i)\in\EE \\
      &\vv_i^+ = \phi^v\big(\vv_i, \rho^{e\to v}\big(\{\ee_{ji}^+\}_{j\in\NN_i}\big), \rho^{\alpha\to v}\big(\{\aa_{jik}^+\}_{(j,k)\in\AA_i}\big), \uu\big),&\forall i\in\VV\\
      &\xx_i^+ = \psi^x(i, \GG_X),&\forall i\in\VV\\
      &\uu^+ = \phi^u\big(\rho^{v\to u}(\{\vv_{i}^+\}_{i\in\VV}), \rho^{e\to u}(\{\ee_{ji}^+\}_{(j,i)\in\EE}), \uu\big)
    \end{alignat*}
  \end{subequations}
  where the global update can contain aggregated information about angles and distances and also the other updates can be generalised as shown above. This architecture is by construction equivariant to transformations in the coordinate embeddings for which both relative distances and angles are preserved.

\section{Obtaining other architectures as special instances}\label{sec:other}
In this appendix we will represent some of the architectures discussed in the main text explicitly as instances of our architecture.

\subsection{Dimenet~\citep{klicpera2020directional}}
Dimenet can be obtained from the AGN by considering edge and distance information in the angle update and using sum aggregation functions as
\begin{subequations}\label{eq:dimenet}
  \begin{alignat*}{3}
  & \aa_{ijk}^+ = \phi^{\alpha} (\ee_{ji}, \|\xx_i-\xx_j\|_2^2,\angle (\xx_i,\xx_j,\xx_k)),\,\,\,\, &&\forall (j,i,k)\in\AA\\
   &\ee_{ji}^+ = \phi^e ( \ee_{ji}, \sum_{i \in \NN_j } \aa_{ijk}^+ ),&&\forall (j,i)\in\EE \\
     &\xx_i^+ = \xx_i ,&&\forall i\in\VV\\
    &\vv_i^+ = \phi^v\big(\vv_i, \sum_{j\in\NN_i}\ee_{ji}^+\big),&&\forall i\in\VV
  \end{alignat*}
\end{subequations}
where $ \|\xx_i-\xx_j\|_2^2$ is defined represented within a set of orthogonal basis functions $e_{RBF}$ and the angles $\angle (\xx_i,\xx_j,\xx_k)$ within a basis defined as $\alpha_{SBF}$.

\subsection{EGCN~\citep{satorras2021en}}
The EGCN network is obtained from the DGN by selecting a specific form for the coordinate update function $\psi^x$ and using the sum aggregation function as $\phi^{e\to v}$, \ie, 
\begin{subequations}\label{eq:egcn}
  \begin{alignat*}{3}
    &\ee_{ji}^+ = \phi^e\big(\vv_j, \vv_i, \ee_{ji}, \|\xx_i-\xx_j\|_2^2\big),\,\,\,\,&&\forall (j,i)\in\EE \\
    &\xx_i^+ = \xx_i + \sum_{j \neq i} (\xx_i-\xx_j) \phi^x(\ee_{ji}),&&\forall i\in\VV\\
    &\vv_i^+ = \phi^v\big(\vv_i, \sum_{j\in\NN_i}\ee_{ji}^+ \big).&&\forall i\in\VV
  \end{alignat*}
\end{subequations}

\subsection{IsoGNN~\citep{horie2021isometric}}
The IsoGNN architecture is defined for tensors of rank-$n$; to compare with the other architectures presented here, we show below the architecture for rank-1 tensors. 
\begin{subequations}\label{eq:isognn}
  \begin{alignat*}{3}
    &\ee_{ji}^+ = \left(\sum_{k, l \in \VV, k \ne l} \mathbf{T}_{ijkl} (\xx_k - \xx_l) \right) \vv_j ,\,\,\,\,&&\forall (j,i)\in\EE \\
  &\xx_i^+ = \xx_i ,&&\forall i\in\VV\\
    &\vv_i^+ = \phi^v\big(\vv_i, \sum_{j\in\NN_i}\ee_{ji}^+\big),&&\forall i\in\VV
  \end{alignat*}
\end{subequations}
where $\mathbf{T}_{ijkl}$ is an untrainable 2-dimensional matrix which is translation and rotation invariant, and determined offline from the data for each class of problem.

\subsection{Other networks}

Standard graph networks can be obtained from ours by skipping some updates or not considering equivariant information. Also other variants, including SchNet~\citep{schutt2017quantum} or TFN~\citep{thomas2018tensor}, can be cast as message passing architectures as shown in~\citep{satorras2021en}.

\section{Additional experiments and implementation details}

\subsection{Common implementation details}\label{sec:common_setup}
The update functions of the networks are all implemented as MLPs. After the graph layers, the produced node embeddings are passed through another MLP, a global pooling layer and a final MLP with output dimension equal to the number of classes for graph classification tasks. For edge classification tasks (TSP), the architecture after the graph layers is a single MLP taking as input source and target nodes and predicting the class of each edge. 
Each network is trained starting from $10$ different initial conditions. The results in the tables contains the mean and standard deviation resulting from the $10$ initialisations.

\subsection{Polytopes classification}\label{sec:poly}
\subsubsection{Specific implementation details}\label{sec:implementation}
All the MLPs have one hidden layer containing $64$ neurons and swish activation function. 
We used $2$ graph layers for AGN and $3$ for both the DGN and GN with embeddings in the hidden layers having dimension $32$.
Aggregation functions and the pooling layer implements mean or sum operations (and are specified in the results' tables).
Adam~\citep{kingma2014method} is used to train the all the models with a learning rate $\alpha=0.001$ and no regularisation for $1000$ epochs. The batch-size is equal to the number of training samples (so $5,6,3$ respectively, for $n=3,4,5$).
\subsubsection{Additional experiments for $n=3$}
\paragraph{Coordinate update~\eqref{eq:eq_map}}
In the main paper we reported results for $n=3$ when the identity function~\eqref{eq:map_I} was used to perform the coordinate update. In Table~\ref{tab:polytope_orth_c} we report the results obtained when using~\eqref{eq:eq_map} for the coordinate updates. As can be seen the main difference is that now SDGN has an accuracy of $0.8$ instead of $0.2$, being unable to correctly classify simplexes and orthoplexes. The other results are qualitatively similar.
\begin{table}[h!]
  \centering
  \tiny
  \begin{tabular}{l|c|c|c|c|c|c|c|}
    & & & \multicolumn{5}{|c|}{test accuracy}  \\\cline{4-8}
    block & $\rho$ & train acc & Orthogonal  & Orthogonal + dilation  & \begin{tabular}{c}Non-orthogonal\\($\mu=0.5$) \end{tabular} & \begin{tabular}{c}Non-orthogonal\\($\mu=1.5$) \end{tabular} & \begin{tabular}{c}Non-orthogonal\\($\mu=3.0$) \end{tabular}\\
    \hline\hline
    AGN & mean      & $1$   & $\bm{1}$            & $\bm{1}$        & $\bm{1}$       & $\bm{1}$  & $\bm{0.96\pm0.04}$ \\
    AGN & sum       & $1$   & $\bm{1}$            & $\bm{1}$        & $\bm{1}$       & $\bm{1}$            & $\bm{1}$            \\
    SDGN & mean     & $0.8$ & $0.8$               & $0.8$           & $0.65\pm 0.05$ & $0.40\pm 0.08$ & $0.36\pm 0.06$ \\
    SDGN & sum      & $1$   & $\bm{1}$            & $\bm{1}$        & $0.89\pm 0.16$ & $0.62\pm 0.24$ & $0.60\pm 0.21$\\
    DGN & mean      & $1$   & $\bm{1}$            & $0.36\pm 0.02$  & $0.25\pm 0.05$ & $0.34\pm 0.02$ & $0.30\pm 0.02$ \\
    DGN & sum       & $1$   & $\bm{1}$            & $0.37\pm 0.04$  & $0.31\pm 0.06$ & $0.34\pm 0.04$ & $0.32\pm 0.05$ \\
    \hline
  \end{tabular}
  \caption{Polytopes classification: training and test accuracy (mean $\pm$ standard deviation over $10$ runs) for $n=3$, for different transformations in the test set.}
  \label{tab:polytope_orth_c}
\end{table}

\paragraph{Adding not all-identical node features}
The experiments run so far used datasets containing only information about node coordinates and the presence of edges. If we add (not-all-identical) node features, then also SDGN with mean aggregation function is able to correctly classify all the polytopes. This happens thanks to the additional features breaking a symmetry making the graphs of the simplex and the orthoplex look identical to the SDGN layer.

\begin{figure}[htbp]
  \centering
  \includegraphics[width=0.7\columnwidth]{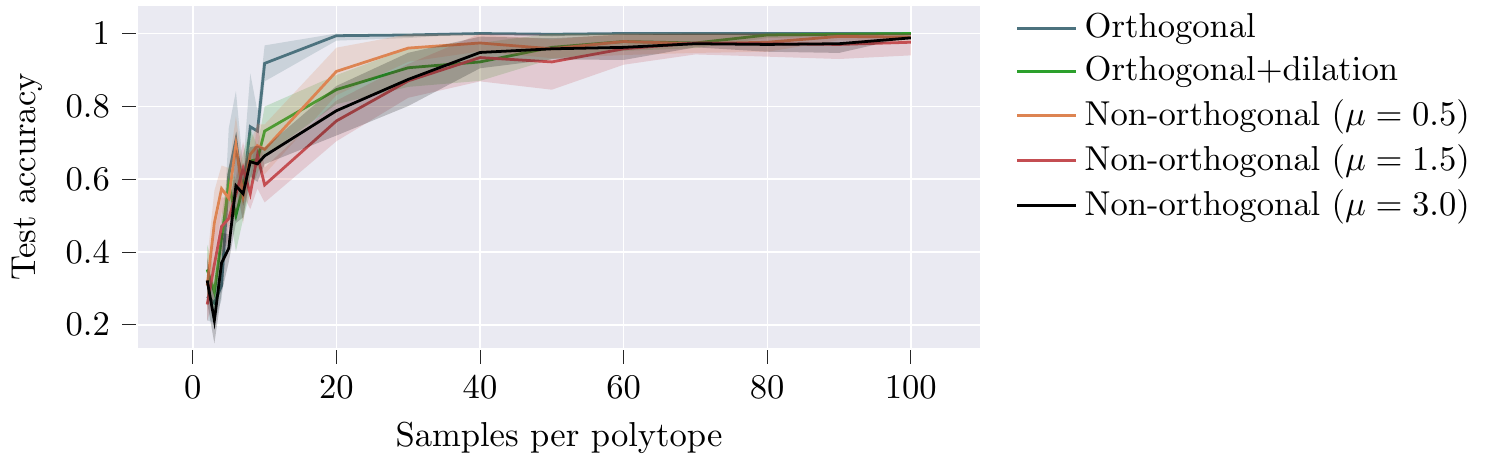}
  \caption{Test accuracy vs samples per polytope in the training set for a standard GNN ($n=3$).}
  \label{fig:efficiency}
\end{figure}

\paragraph{Data efficiency}
To emphasise the advantage of having a network that is able to exploit symmetries in the dataset in terms of data efficiency, we study how many samples in the training set are necessary for a standard GNN to reach reasonable generalisation performance. For the set of transformations we considered in the previous sections, we augmented the training set with $\{2,3,\dots,100\}$ randomly transformed (as in the respective test set) copies of each polytope. We trained a standard GNN on these augmented datasets and observed the resulting test accuracy after $1000$ epochs.
Results are reported in Figure~\ref{fig:efficiency} for each set of transformations in terms of mean and standard deviation over $10$ random initialisations. It can be seen that $20$ samples per polytope may be sufficient when only orthogonal transformations are considered. Adding also dilations and non-orthogonal transformations further increases the number of data points that are required.
This shows that while data augmentation can be successfully exploited, it is provably sub-optimal in terms of sample complexity~\citep{mei2021learning} and architectures with built-in equivariance properties represent a more efficient strategy to consider.

\subsubsection{Experiments in higher dimensions}
For $n=4$ simplexes, hypercubes, orthoplexes are considered along with $24$-, $120$- and $600$-cell polytopes.
For $n>5$ only simplexes, hypercubes, orthoplexes are considered.

In Tables~\ref{tab:polytope_n4},~\ref{tab:polytope_n5} we report the results obtained for $n=4$ and $n=5$. Similarly as before, also for $n=4,5$ there are cases in which the DGN (or SDGN) is not able to distinguish between one or more polytopes. Moreover, for $n=5$ we note that SDGN with sum aggregation and $\phi^x$ as in~\eqref{eq:map_I} perfectly generalise to unseen data subject to non-orthogonal transformations, while the AGN with mean aggregation function has a drop in performance as $\mu$ increases.

\begin{table}[h!]
  \centering
  \tiny
  \begin{tabular}{l|c|c|c|c|c|c|c|c|}
    & & & & \multicolumn{5}{|c|}{test accuracy}  \\\cline{5-9}
    block & $\rho$ & $\psi^x$ & train acc & Orthogonal  & Orthogonal + dilation  & \begin{tabular}{c}Non-orthogonal\\($\mu=0.5$) \end{tabular} & \begin{tabular}{c}Non-orthogonal\\($\mu=1.5$) \end{tabular} & \begin{tabular}{c}Non-orthogonal\\($\mu=3.0$) \end{tabular}\\
    \hline\hline
    AGN & mean   & \eqref{eq:map_I}   & $1$           & $\bm{1}$        & $\bm{1}$        & $\bm{1}$       & $\bm{1}$       & $\bm{0.98\pm0.04}$ \\
    AGN & mean   & \eqref{eq:eq_map}  & $1$           & $\bm{1}$        & $\bm{1}$        & $\bm{1}$       & $\bm{1}$       & $\bm{0.97\pm0.04}$ \\
    AGN & sum    & \eqref{eq:map_I}   & $1$           & $\bm{1}$        & $\bm{1}$        & $\bm{1}$       & $\bm{1}$       & $\bm{1}$            \\
    AGN & sum    & \eqref{eq:eq_map}  & $1$           & $\bm{1}$        & $\bm{1}$        & $\bm{1}$       & $\bm{1}$       & $\bm{1}$            \\
    SDGN & mean  & \eqref{eq:map_I}   & $0.17$        & $0.17$          & $0.17$          & $0.17$         & $0.17$         & $0.17$ \\
    SDGN & mean  & \eqref{eq:eq_map}  & $1$           & $\bm{1}$        & $\bm{1}$        & $0.57\pm 0.13$ & $0.35\pm 0.10$ & $0.27\pm 0.10$ \\
    SDGN & sum   & \eqref{eq:map_I}   & $1$           & $\bm{1}$        & $\bm{1}$        & $0.96\pm 0.03$ & $0.85\pm 0.13$ & $0.77\pm 0.17$\\
    SDGN & sum   & \eqref{eq:eq_map}  & $1$           & $\bm{1}$        & $\bm{1}$        & $0.82\pm 0.11$ & $0.68\pm 0.14$ & $0.65\pm 0.20$\\
    DGN & mean   & \eqref{eq:map_I}   & $0.83$        & $0.83$          & $0.31\pm 0.02$  & $0.40\pm 0.03$ & $0.32\pm 0.01$ & $0.27\pm 0.04$ \\
    DGN & mean   & \eqref{eq:eq_map}  & $1$           & $\bm{1}$        & $0.34\pm 0.02$  & $0.40\pm 0.04$ & $0.38\pm 0.03$ & $0.30\pm 0.03$ \\
    DGN & sum    & \eqref{eq:map_I}   & $1$           & $\bm{1}$        & $0.50\pm 0.05$  & $0.54\pm 0.08$ & $0.45\pm 0.06$ & $0.39\pm 0.08$ \\
    DGN & sum    & \eqref{eq:eq_map}  & $1$           & $\bm{1}$        & $0.50\pm 0.04$  & $0.54\pm 0.04$ & $0.52\pm 0.03$ & $0.43\pm 0.06$ \\
    GN & mean    & $-$                & $1$           & $0.14\pm 0.04$  & $0.19\pm 0.05$  & $0.21\pm 0.03$ & $0.24\pm 0.02$ & $0.22\pm 0.04$ \\
    GN & sum     & $-$                & $1$           & $0.31\pm 0.07$  & $0.31\pm 0.06$  & $0.31\pm 0.05$ & $0.33\pm 0.06$ & $0.23\pm 0.04$ \\
    \hline
  \end{tabular}
  \caption{Polytopes classification: training and test accuracy (mean $\pm$ standard deviation over $10$ runs) for $n=4$, for different transformations in the test set.}
  \label{tab:polytope_n4}
\end{table}

\begin{table}[h!]
  \centering
  \tiny
  \begin{tabular}{l|c|c|c|c|c|c|c|c|}
    & & & & \multicolumn{5}{|c|}{test accuracy}  \\\cline{5-9}
    block & $\rho$ & $\psi^x$ & train acc & Orthogonal  & Orthogonal + dilation  & \begin{tabular}{c}Non-orthogonal\\($\mu=0.5$) \end{tabular} & \begin{tabular}{c}Non-orthogonal\\($\mu=1.5$) \end{tabular} & \begin{tabular}{c}Non-orthogonal\\($\mu=3.0$) \end{tabular}\\
    \hline\hline
    AGN & mean   & \eqref{eq:map_I}   & $1$           & $\bm{1}$        & $\bm{1}$        & $\bm{0.99\pm 0.03}$ & $0.86\pm 0.11$ & $0.77\pm 0.13$ \\
    AGN & mean   & \eqref{eq:eq_map}  & $1$           & $\bm{1}$        & $\bm{1}$        & $\bm{0.99\pm 0.03}$ & $0.84\pm 0.11$ & $0.75\pm 0.11$ \\
    AGN & sum    & \eqref{eq:map_I}   & $1$           & $\bm{1}$        & $\bm{1}$        & $\bm{1}$            & $\bm{1}$       & $\bm{1}$            \\
    AGN & sum    & \eqref{eq:eq_map}  & $1$           & $\bm{1}$        & $\bm{1}$        & $\bm{1}$            & $\bm{1}$       & $\bm{1}$            \\
    SDGN & mean  & \eqref{eq:map_I}   & $0.33$        & $0.31\pm 0.08$  & $0.31\pm 0.08$  & $0.33$              & $0.33$         & $0.33$ \\
    SDGN & mean  & \eqref{eq:eq_map}  & $1$           & $\bm{1}$        & $\bm{1}$        & $0.60\pm 0.25$      & $0.48\pm 0.15$ & $0.37\pm 0.10$ \\
    SDGN & sum   & \eqref{eq:map_I}   & $1$           & $\bm{1}$        & $\bm{1}$        & $\bm{1}$            & $\bm{1}$       & $\bm{1}$ \\
    SDGN & sum   & \eqref{eq:eq_map}  & $1$           & $\bm{1}$        & $\bm{1}$        & $0.83\pm 0.16$      & $0.75\pm 0.14$ & $0.66\pm 0.18$\\
    DGN & mean   & \eqref{eq:map_I}   & $1$           & $\bm{1}$        & $0.49\pm 0.01$  & $0.33\pm 0.00$      & $0.49\pm 0.02$ & $0.37\pm 0.01$ \\
    DGN & mean   & \eqref{eq:eq_map}  & $1$           & $\bm{1}$        & $0.51\pm 0.03$  & $0.40\pm 0.04$      & $0.41\pm 0.06$ & $0.39\pm 0.05$ \\
    DGN & sum    & \eqref{eq:map_I}   & $1$           & $\bm{1}$        & $0.53\pm 0.05$  & $0.34\pm 0.05$      & $0.51\pm 0.03$ & $0.41\pm 0.10$ \\
    DGN & sum    & \eqref{eq:eq_map}  & $1$           & $\bm{1}$        & $0.57\pm 0.10$  & $0.45\pm 0.12$      & $0.54\pm 0.13$ & $0.49\pm 0.10$ \\
    GN & mean    & $-$                & $1$           & $0.30\pm 0.05$  & $0.52\pm 0.08$  & $0.34\pm 0.04$      & $0.38\pm 0.08$ & $0.43\pm 0.06$ \\
    GN & sum     & $-$                & $1$           & $0.39\pm 0.07$  & $0.49\pm 0.04$  & $0.34\pm 0.08$      & $0.40\pm 0.06$ & $0.41\pm 0.08$ \\
    \hline
  \end{tabular}
  \caption{Polytopes classification: training and test accuracy (mean $\pm$ standard deviation over $10$ runs) for $n=5$, for different transformations in the test set.}
  \label{tab:polytope_n5}
\end{table}

\subsection{MNIST, CIFAR10, TSP}
MNIST and CIFAR10 are classical image classification datasets converted into graphs using super-pixels and assigning the super-pixel coordinates as node coordinates and the intensity as node features. TSP (based on the Travelling Salesman Problem) tests link prediction on 2D Euclidean graphs to identify edges belonging to the optimal TSP solution given by the Concorde solver.

For this datasets we use the same splits as specified in~\citep{dwivedi2020benchmarking}.

\subsubsection{Specific implementation details}
The MLPs have one hidden layer containing $64$ neurons, swish activation function and dropout layers with dropout rate $0.01$.
We used $2$ graph layers for AGN and $3$ for both the DGN and GN with embeddings in the hidden layers having dimension $64$.
Aggregation functions and the pooling layer implements sum operations.
The models are trained with Adam with learning rate $\alpha=0.001$ and no regularisation for $100$ epochs. Batch-size $128$ is used for MNIST and CIFAR10, and $8$ for TSP.

\subsection{Molecular property prediction - QM9} \label{sec:qm9_results}
The QM9 dataset~\citep{DBLP:journals/corr/WuRFGGPLP17} is comprised of small molecules (hydrogen, carbon, nitrogen, oxygen, flourine) with the target properties being 12 chemical properties. As the target properties are equivariant to Euclidean transformations of the atoms' coordinates, and also to the order in which atoms are processed, QM9 is an excellent benchmarking dataset for a GNN, especially if $E(n)$ equivariant. Indeed, state of the art results have been achieved on this dataset by EGCN and Dimenet~\citep{satorras2021en,klicpera2020directional}. Here we show that, in addition to leading to better accuracy, employing equivariant networks give a significant improvement in convergence speed.

\begin{figure}
  \centering
  \begin{subfigure}{0.24\linewidth}
    \centering
    \includegraphics[width=\linewidth]{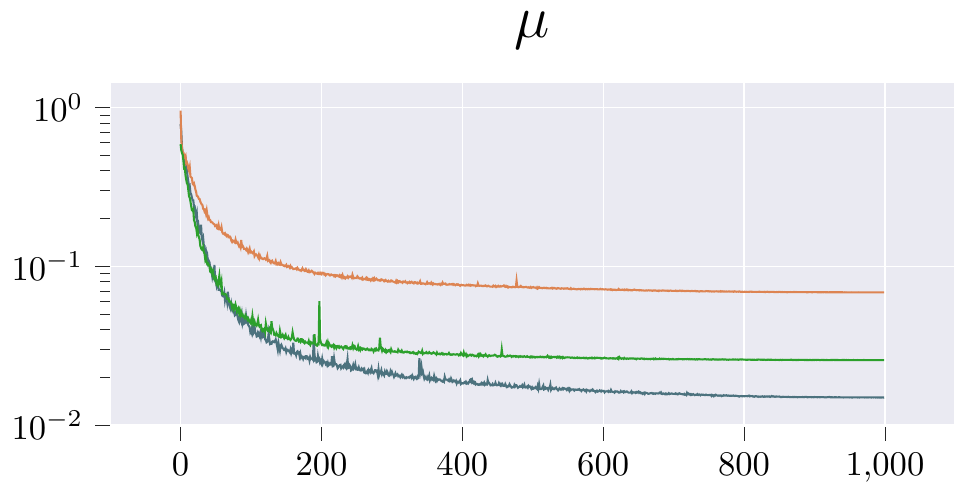}
  \end{subfigure}
  \begin{subfigure}{0.24\linewidth}
    \centering
    \includegraphics[width=\linewidth]{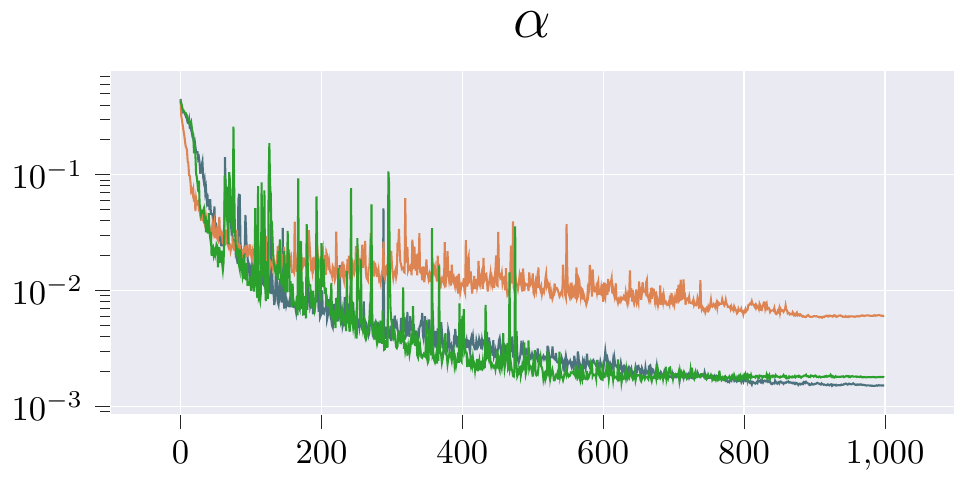}
  \end{subfigure}
  \begin{subfigure}{0.24\linewidth}
    \centering
    \includegraphics[width=\linewidth]{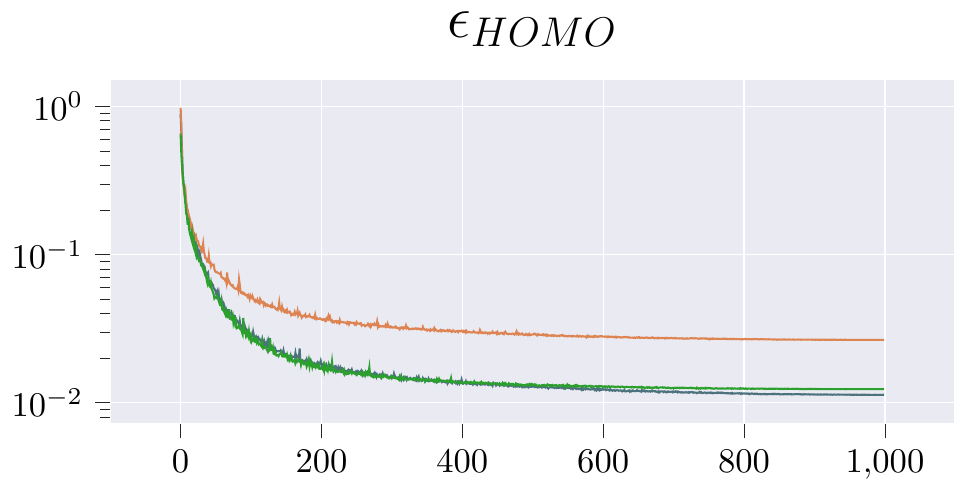}
  \end{subfigure}
  \begin{subfigure}{0.24\linewidth}
    \centering
    \includegraphics[width=\linewidth]{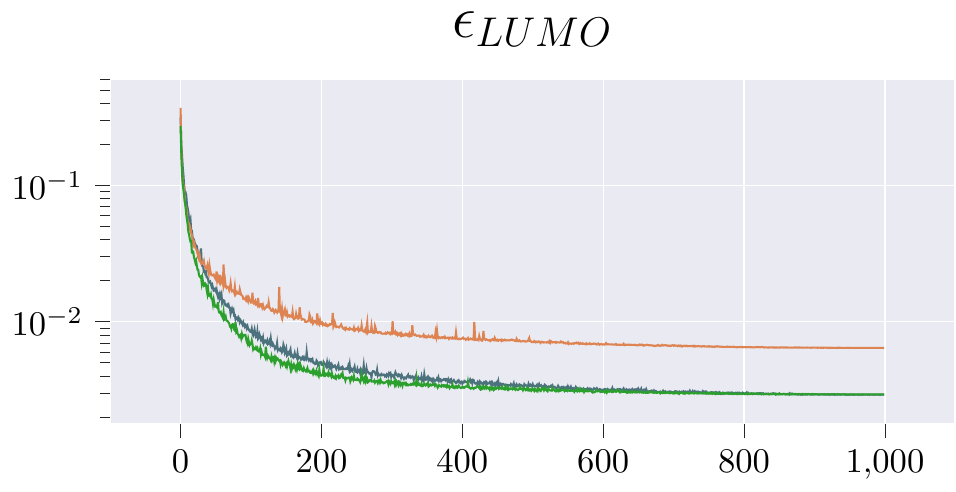}
  \end{subfigure}

  \begin{subfigure}{0.24\linewidth}
    \centering
    \includegraphics[width=\linewidth]{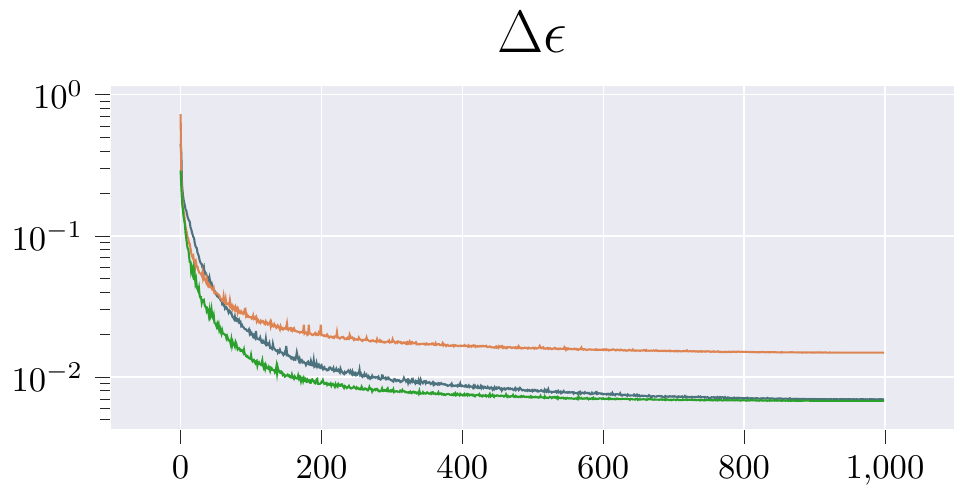}
  \end{subfigure}
  \begin{subfigure}{0.24\linewidth}
    \centering
    \includegraphics[width=\linewidth]{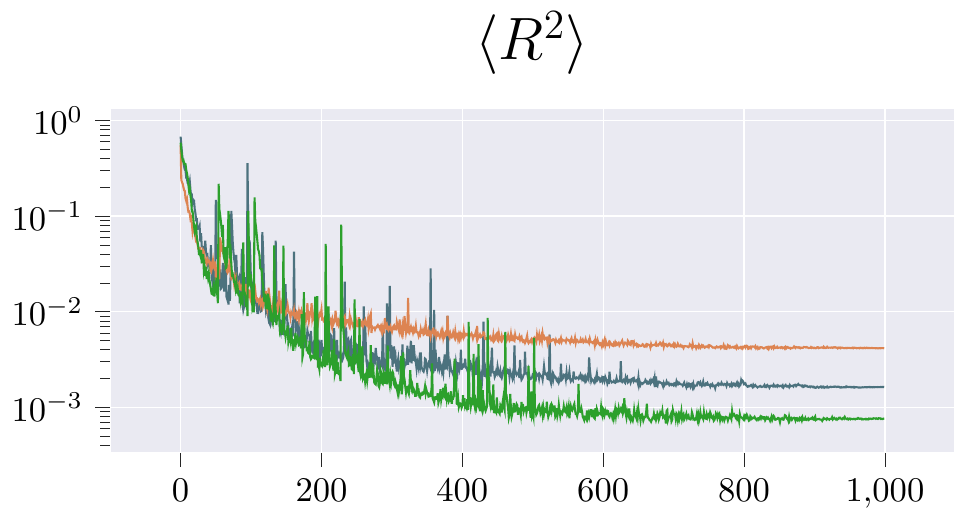}
  \end{subfigure}
  \begin{subfigure}{0.24\linewidth}
    \centering
    \includegraphics[width=\linewidth]{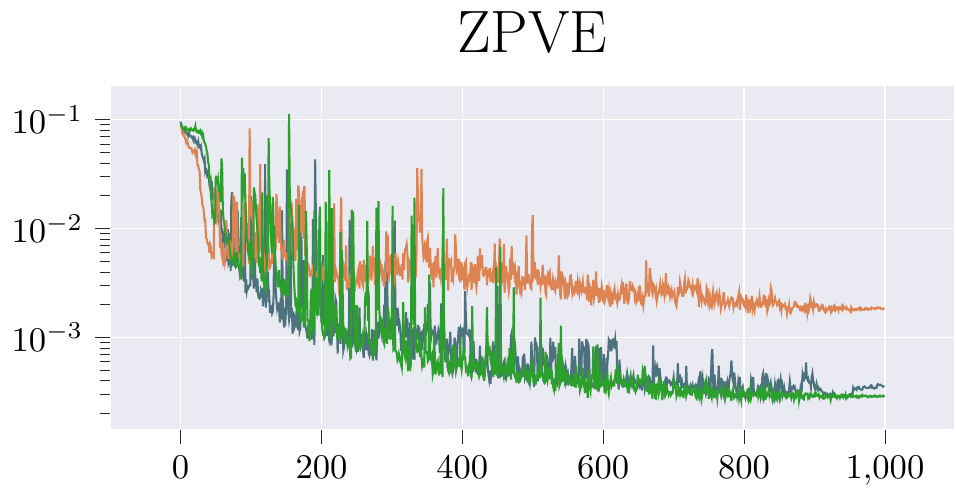}
  \end{subfigure}
  \begin{subfigure}{0.24\linewidth}
    \centering
    \includegraphics[width=\linewidth]{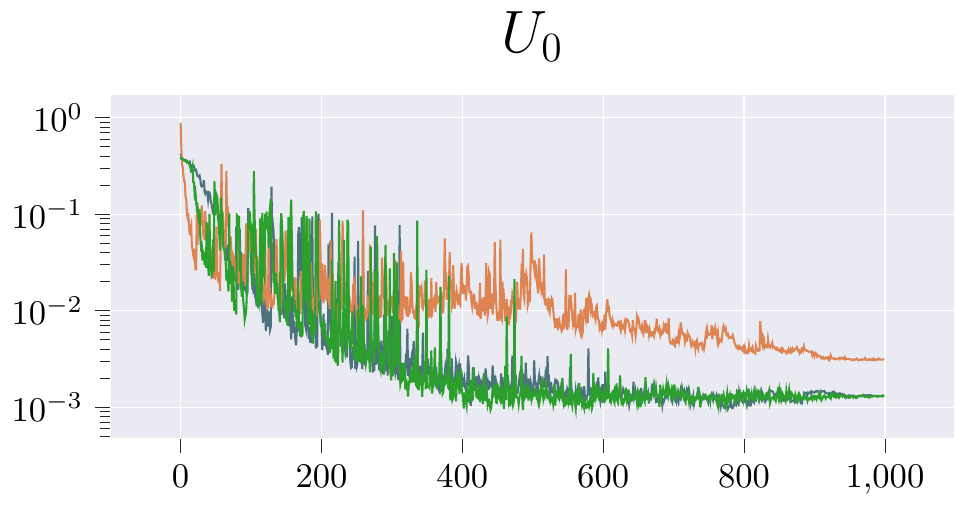}
  \end{subfigure}

  \begin{subfigure}{0.24\linewidth}
    \centering
    \includegraphics[width=\linewidth]{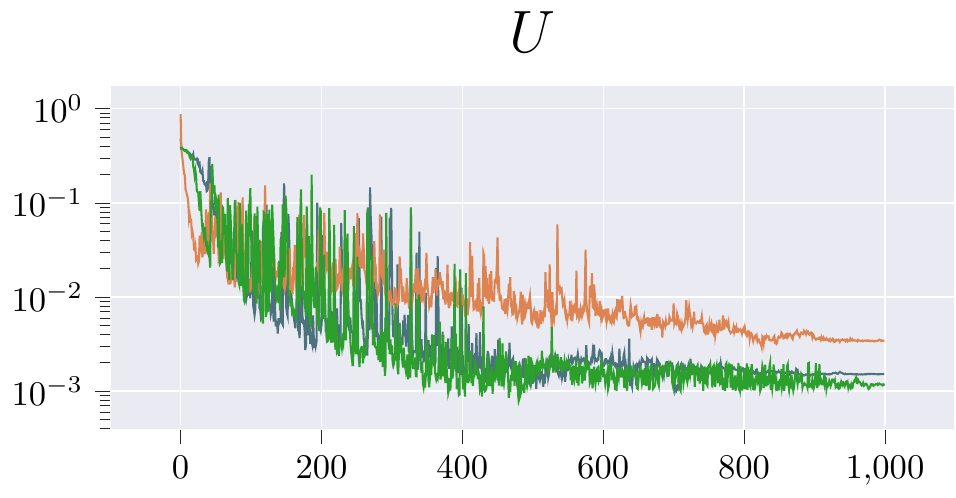}
  \end{subfigure}
  \begin{subfigure}{0.24\linewidth}
    \centering
    \includegraphics[width=\linewidth]{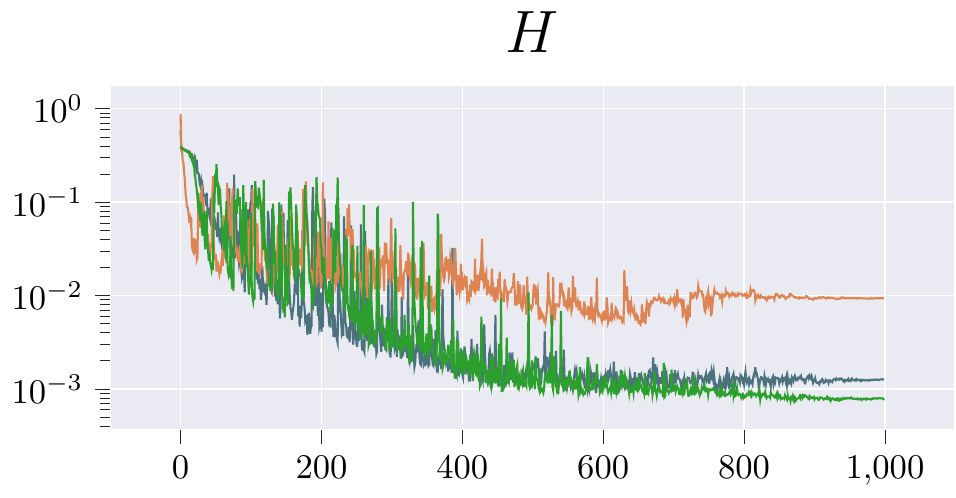}
  \end{subfigure}
  \begin{subfigure}{0.24\linewidth}
    \centering
    \includegraphics[width=\linewidth]{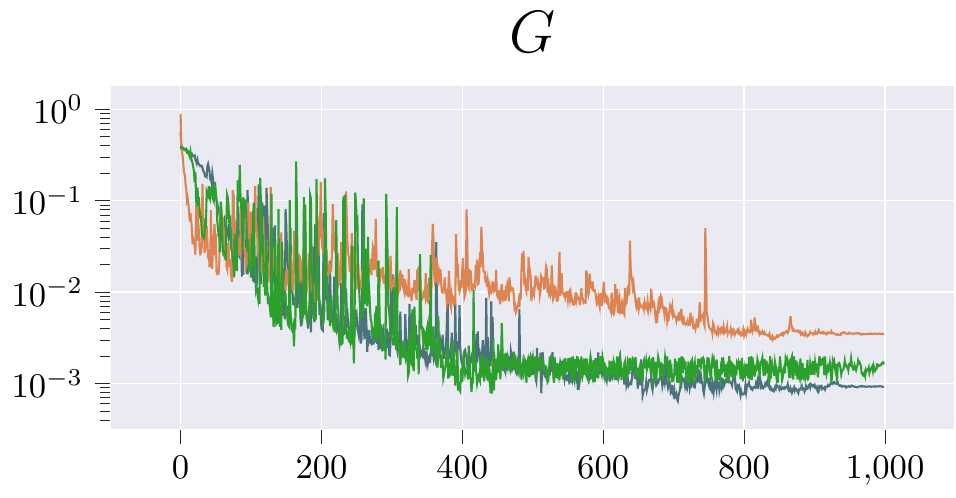}
  \end{subfigure}
  \begin{subfigure}{0.24\linewidth}
    \centering
    \includegraphics[width=\linewidth]{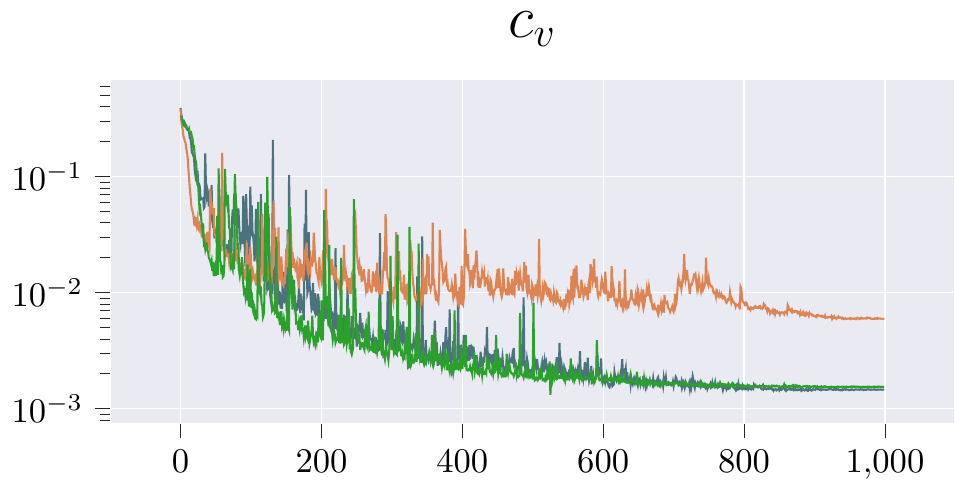}
  \end{subfigure}

  \vspace{2ex}

  \includegraphics[width=0.3\linewidth]{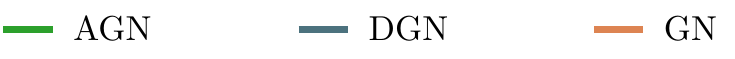}
  \caption{QM9: test MSE loss on the 12 target properties.}
  \label{fig:qm9_mse}
\end{figure}

Figure~\ref{fig:qm9_mse} show the evolution of the mean squared error on the test set on all the target properties for both architectures. We see that both the AGN and DGN outperform the standard GN in two aspects; the model trains more rapidly, and also reaches a lower loss. 

\subsubsection{Specific implementation details}
The QM9 dataset is composed of roughly 134k molecules: we used 100k for training and the remaining for testing. The AGN and DGN networks receive the embedding of the atomic properties as initial node features and the coordinates of each atom as coordinate features, while in the standard GNN they are stacked and provided as node features. Edge embeddings representing bond types are also provided. 

The networks consist of 4 graph layers (3 for the AGN) with each MLP having 2 hidden layers of 128 nodes, swish activation function and dropout rate of 0.01. The node embeddings of the last graph layer are passed through an MLP, a global mean pooling layer and a last MLP to map them to the target chemical property. The last two MLPs each have a single hidden layer of 128 nodes, swish activation and dropout rate of 0.01. The target chemical properties are all standardised by subtracting the mean and dividing by the standard deviation for each target. The networks are trained for 1000 epochs using ADAM with a learning rate of $0.0005$ and mean squared error loss.

\end{document}